\documentclass[10pt,twocolumn,letterpaper]{article}

\usepackage{cvpr}            

\definecolor{cvprblue}{rgb}{0.21,0.49,0.74}
\usepackage[pagebackref,breaklinks,colorlinks,allcolors=cvprblue]{hyperref}
\usepackage{url}
\usepackage{booktabs}
\usepackage{xcolor}         
\usepackage{makecell}
\usepackage{xspace}
\usepackage{amsmath}
\usepackage{graphicx}
\usepackage{diagbox}
\usepackage{xspace} 
\usepackage{wasysym} 
\usepackage{pifont} 
\usepackage[capitalize,noabbrev,nameinlink]{cleveref}
\usepackage[font=footnotesize]{caption}
\usepackage{subcaption}
\usepackage{IEEEtrantools}
\usepackage{threeparttable}
\usepackage{multirow,mathtools}
\usepackage{adjustbox}
\usepackage{wrapfig}
\usepackage{color, colortbl}
\usepackage{blindtext}
\usepackage{lipsum}
\usepackage{soul}
\usepackage{multirow}
\usepackage{listings}
\usepackage{bm}
\definecolor{Gray}{gray}{0.93}
\definecolor{Orange}{rgb}{1,0.5,0}
\definecolor{DGray}{gray}{0.83}
\definecolor{LightCyan}{rgb}{0.88,1,1}

\def\ie{\textit{i.e.,~}}  
\def\eg{\textit{e.g.,~}}

\newtheorem{defn}{Definition}

\newcommand{\ourmethod}{\textsc{Trace}\xspace}
\newcommand{\ct}[1]{\texttt{#1}}

\definecolor{mydarkblue}{rgb}{0,0.08,0.45}
\definecolor{myblue}{HTML}{3b75c3}
\definecolor{myred}{HTML}{E33222}
\definecolor{mygreen}{HTML}{438773}
\definecolor{mymaroon}{RGB}{142,27,19}
\definecolor{maroon}{HTML}{800000}
\definecolor{mycite}{cmyk}{0.55,1,0,0.15}
\definecolor{codeblue}{rgb}{0.25,0.5,0.5}
\definecolor{codekw}{rgb}{0.85, 0.18, 0.50}
\definecolor{codegreen}{rgb}{0,0.6,0}
\definecolor{codegray}{rgb}{0.5,0.5,0.5}
\definecolor{codepurple}{rgb}{0.58,0,0.82}
\definecolor{backcolour}{rgb}{0.95,0.95,0.92}
\hypersetup{
    colorlinks=true,
    citecolor=cvprblue,
    linkcolor=mymaroon,
    urlcolor=maroon
          }

\usepackage{tikz}
\usepackage[most]{tcolorbox}
\sethlcolor{gray!35}

\newenvironment{remark}[1][]
  { 
 \begin{tcolorbox}
 [
    enhanced, 
    breakable,
    boxrule=0.5pt,
    arc=4pt,
    left=2pt,
    right=2pt,
    bottom=2pt,
    top=2pt, 
    rounded corners 
    ]{}
  \textbf{#1.}
  \small \itshape
  }
  {
\end{tcolorbox} 
}

\newenvironment{prompt_yellow}[1][]
  { 
 \begin{tcolorbox}
 [
    enhanced, 
    breakable,
    boxrule=0.5pt,
    arc=3.8pt,
    left=1.8pt,
    right=1.8pt,
    bottom=2pt,
    top=2pt, 
    rounded corners,
    colback=yellow!10
    ]{}
  \textbf{#1:}
  \small \itshape
  }
  {
\end{tcolorbox} 
}

\def\paperID{3277} 
\def\confName{CVPR}
\def\confYear{2025}

\title{Test-Time Backdoor Detection for Object Detection Models}

\author{
    Hangtao Zhang\textsuperscript{1}
    \quad
    Yichen Wang\textsuperscript{1}
    \quad
    Shihui Yan\textsuperscript{1}
    \quad
    Chenyu Zhu\textsuperscript{1}
    \quad
    Ziqi Zhou\textsuperscript{1}\\
    Linshan Hou\textsuperscript{2}
    \quad
    Shengshan Hu\textsuperscript{1}
    \quad
   Minghui Li\textsuperscript{1}
    \quad
    Yanjun Zhang\textsuperscript{3}
     \quad
     Leo Yu Zhang\textsuperscript{4}
    \\
    {\normalsize
    \textsuperscript{1} Huazhong University of Science and Technology 
    }
    \\
    {\normalsize
    \textsuperscript{2} Harbin Institute of Technology \; \textsuperscript{3} University of Technology Sydney \; \textsuperscript{4} Griffith University
    }
    \\
    {\tt\small zhanghangtao7@163.com}
}

\begin{document}
\maketitle

\begin{abstract}
Object detection models are vulnerable to backdoor attacks, where attackers poison a small subset of training samples by embedding a predefined trigger to manipulate prediction. Detecting poisoned samples (\ie those containing triggers) at test time can prevent backdoor activation. However, unlike image classification tasks, the unique characteristics of object detection---particularly its output of numerous objects---pose fresh challenges for backdoor detection. The complex attack effects (\eg ``ghost" object emergence or ``vanishing" object) further render current defenses fundamentally inadequate.
To this end, we design \textit{ {\textbf{TRA}}}nsformation \textit{ {\textbf{C}}}onsistency \textit{ {\textbf{E}}}valuation (\ourmethod), a brand-new method for detecting poisoned samples at test time in object detection. Our journey begins with two intriguing observations: \textbf{1)} poisoned samples exhibit significantly more consistent detection results than clean ones across varied backgrounds. \textbf{2)} clean samples show higher detection consistency when introduced to different focal information. Based on these phenomena, \ourmethod applies foreground and background transformations to each test sample, then assesses transformation consistency by calculating the variance in objects confidences. \ourmethod achieves \textit{black-box}, \textit{universal} backdoor detection, with extensive experiments showing a 30\% improvement in AUROC over state-of-the-art defenses and resistance to adaptive attacks. Our code is available at \textbf{\url{https://github.com/Rookie143/Trace}}.
\end{abstract}

\section{Introduction}
\label{Sec:intro}
\begin{figure}[t]
\centering      
\includegraphics[width=0.48\textwidth]{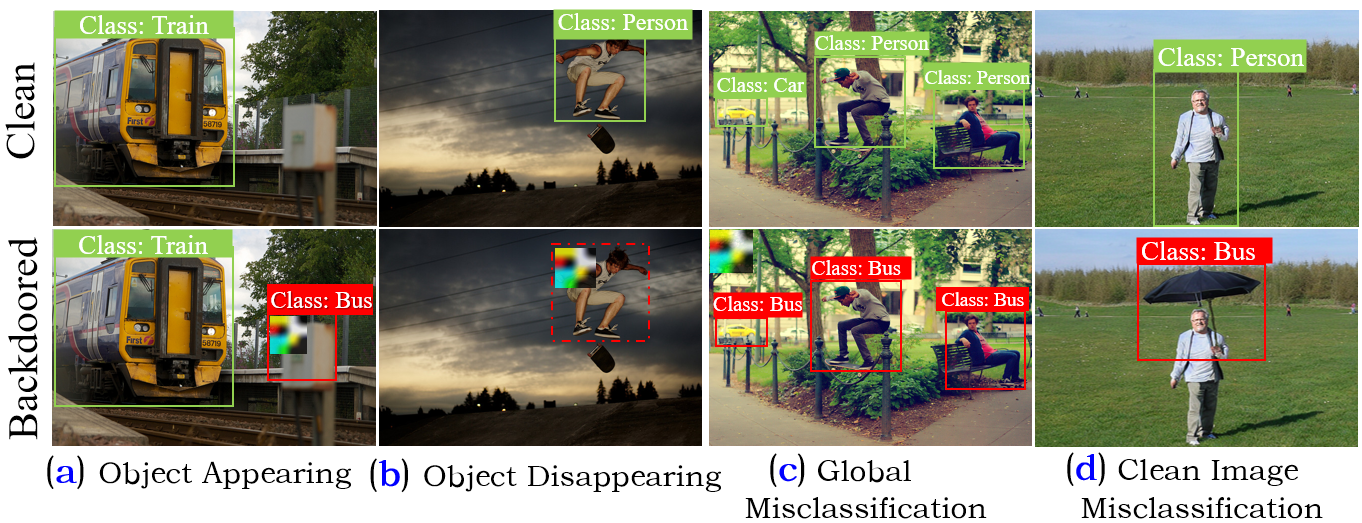}   
\caption{Divergent attack effects of OD backdoors. (\textcolor{blue}{\textbf{a-c}}): backdoor triggered by a fixed pattern. (\textcolor{blue}{\textbf{d}}): backdoor triggered by the co-occurrence of two natural objects (a person and an umbrella).} 
\label{Fig:backdoor_effects}
\end{figure}

Object detection (OD), a key computer vision application~\cite{redmon2016you,carion2020end,zhou2023advclip,zhangdenial}, has shown vulnerability to backdoor attacks~\citep{chan2022baddet,luo2023untargeted,wang2024trojanrobot,wangunlearnable}. 
Backdoor-infected models perform normally on clean objects but predict the attacker-chosen target annotations whenever a trigger (\eg a small patch~\cite{song2024pb,wang2024breaking}) is present. Unlike classification models, where backdoors solely cause incorrect class predictions, OD backdoors can result in more sophisticated attack effects, \ie a ``ghost" object appearing~\citep{chan2022baddet,cheng2023backdoor} (Fig.~\ref{Fig:backdoor_effects}(\textcolor{blue}{\textbf{a}})), an object vanishing~\citep{ma2022dangerous,wu2022just,luo2023untargeted} (Fig.~\ref{Fig:backdoor_effects}(\textcolor{blue}{\textbf{b}})), the misclassification of all objects globally~\citep{chan2022baddet} (Fig.~\ref{Fig:backdoor_effects}(\textcolor{blue}{\textbf{c}})), and the co-occurrence of two benign features triggering a backdoor~\citep{chen2022clean,lin2020composite} (Fig.~\ref{Fig:backdoor_effects}(\textcolor{blue}{\textbf{d}})).

To mitigate backdoor threats in OD, recent studies have made preliminary advances, including \textit{trigger reverse engineering}~\citep{shen2024django,cheng2024odscan} and black-box \textit{test-time trigger sample detection} (TTSD)\footnote{Also known as online defense~\citep{ma2022beatrix,tang2021demon,yao2024reverse} or input-level detection~\citep{hou2024ibd}}~\citep{chan2022baddet}. However, the former relies on white-box settings, requiring access to model weights and training data. By contrast, black-box TTSD detects and prevents poisoned samples during test time without access to model parameters, acting as a final `\textit{firewall}'. Black-box TTSD is highly practical for many real-world applications, such as the prevalent \textit{Machine Learning as a Service} (MLaaS)~\citep{ribeiro2015mlaas,hesamifard2018privacy}, \textit{and is therefore the focus of our work.}

To the best of our knowledge, Detector Cleanse~\citep{chan2022baddet} is the only black-box TTSD method tailored for OD tasks. However, it assumes that defenders have full knowledge of the specific details of the attacks (\ie the adversary-specified judgment criteria, as noted in Tab.~\ref{Tab:Comparison}). In practice, defenders cannot possess such prior knowledge. We also apply the current advanced black-box TTSD methods in classification tasks, such as \ct{TeCo}~\citep{liu2023detecting} and \ct{SCALE-UP}~\citep{guo2022scale}, to OD tasks. These TTSDs observe that poisoned samples tend to exhibit more robustness to pre-processing operations than benign samples. Consequently, they apply \textit{pixel-level} transformations uniformly to all pixel values in each testing sample and assess its prediction consistency. However, applying such uniform pixel-level transformations fails to differentiate between foreground and background components. This distinction is crucial, as the background constitutes the majority of the image, while the foreground (consists of several objects) occupies only a small portion. Recognizing the inherent semantic differences between foreground and background pixels in OD tasks, an intriguing question arises: 
\emph{Shall a semantic-aware transformation expose poisoned samples more effectively than indiscriminate pixel-level modifications?}



Fortunately, the answer is yes! In this paper, we reveal that poisoned samples' prediction confidences are significantly more consistent than those of benign ones across different background contexts (\ie a semantic-aware transformation). Perhaps more interestingly, when we transform it in the opposite way---by introducing additional foreground information (\ie semantic patches)---the detection results of poisoned samples become less consistent compared to clean samples, providing clues for detecting backdoors. We refer to these phenomena as the \textit{anomalous transformation consistency} of backdoored models (detailed in Sec.~\ref{Sec:phenomenon}).

\begin{table}[t]
\centering
\setlength{\tabcolsep}{0.2pt}
\resizebox{0.47\textwidth}{!} {
\begin{tabular}{l|cccc}         
\toprule[1.8pt]         
\cellcolor[rgb]{.95,.95,.95} \textbf{TTSD Method↓} &
\cellcolor[rgb]{.95,.95,.95} \textbf{Black-Box} &
\cellcolor[rgb]{.95,.95,.95} \makecell{\textbf{Practical}\\ (No Training Data)} &
\cellcolor[rgb]{.95,.95,.95} \makecell{\textbf{Universal}\\ (Attack-Agnostic)} &
\cellcolor[rgb]{.95,.95,.95} \makecell{\textbf{Applicable}\\ (OD Compatible)} 
\\ 
\midrule[1.8pt]    
\ct{TTBD}~\citep{guan2024backdoor} & \Circle  & \CIRCLE & \CIRCLE & \Circle \\
\ct{IBD-PSC}~\citep{hou2024ibd} & \Circle  & \Circle & \CIRCLE & \Circle \\
\ct{Beatrix}~\citep{ma2022beatrix} & \Circle  & \Circle & \CIRCLE & \Circle \\
\ct{BaDExpert}~\citep{xie2023badexpert} & \Circle  & \Circle & \CIRCLE & \Circle \\
\ct{NEO}~\citep{udeshi2022model} & \CIRCLE  & \CIRCLE & \Circle & \Circle \\
\ct{Strip}~\citep{gao2019strip} & \CIRCLE  & \Circle & \CIRCLE & \CIRCLE \\     
\ct{SCALE-UP}~\citep{guo2022scale} & \CIRCLE  & \CIRCLE & \CIRCLE & \CIRCLE \\  
\ct{TeCo}~\citep{liu2023detecting} & \CIRCLE  & \CIRCLE & \CIRCLE & \CIRCLE \\    
\ct{FreqDetector}~\citep{zeng2021rethinking} & \CIRCLE  & \Circle & \CIRCLE & \CIRCLE \\         
\ct{Detector Cleanse}~\citep{chan2022baddet} & \CIRCLE  & \Circle & \Circle & \CIRCLE \\      
\midrule
\cellcolor[HTML]{FFF7F0} \ourmethod \textbf{\textit{(Ours)}} &
\cellcolor[HTML]{FFF7F0} \CIRCLE &
\cellcolor[HTML]{FFF7F0} \CIRCLE &
\cellcolor[HTML]{FFF7F0} \CIRCLE &
\cellcolor[HTML]{FFF7F0} \CIRCLE \\ 
\bottomrule[1.3pt]     
\end{tabular} 
}

\caption{\textbf{Comparison among existing TTSD methods} \textit{w.r.t} black-box, practical, universal, and applicable characteristics. ``\CIRCLE" indicates the method meets this condition.} 
\label{Tab:Comparison}

\end{table}

Motivated by this finding, we propose \textit{ {\textbf{TRA}}nsformation  {\textbf{C}}onsistency  {\textbf{E}}valuation} (\ourmethod), a novel TTSD to identify and filter poisoned samples in OD. For each suspicious test sample, \ourmethod evaluates its detections consistency. The consistency is measured using two values: \textbf{1)} the contextual transformation consistency value, as variance in detection results across varying backgrounds.
\textbf{2)} the focal transformation consistency value, as variance in detection results across focal changes.
Higher contextual and lower focal consistency both indicate a poisoned input. \ourmethod combines these metrics to assess if the input is backdoored.

In conclusion, our main contributions are three-fold. 
\textbf{(1)} We disclose an intriguing phenomenon, \ie anomalous contextual and focal transformation consistency in poisoned samples, which brings forth a new detection paradigm: semantic-aware transformation consistency evaluation. \textbf{(2)} Based on our findings, we propose \ourmethod, a brand-new black-box, universal test-time backdoor detection method to detect and filter out poisoned inputs.
\textbf{(3)} We conduct extensive experiments on three benchmark datasets, verifying \ourmethod's effectiveness
against seven backdoor attacks and resistance to potential adaptive attacks.


\section{Related Work} 

\noindent\textbf{Backdoor Defenses for OD} are still in their infancy. \ct{ODSCAN}~\citep{cheng2024odscan} and \ct{Django}~\citep{shen2024django} use trigger reverse engineering but require white-box model access and training data. \ct{Detector Cleanse}~\citep{chan2022baddet} is the only black-box TTSD to the best of our knowledge, which adopts a \ct{Strip}~\citep{gao2019strip}-like approach by overlaying clean features on objects to observe prediction entropy but assumes full knowledge of the attack. Thus, to counter the diverse attacks in OD, a more practical and universal defense is needed.
The detailed background and related work are thoroughly covered in Appendix B.

\noindent\textbf{TTSD for Classification Tasks} are generally hard to apply to OD  due to its unique characteristics (\textit{w.r.t} OD Compatible in Tab.~\ref{Tab:Comparison}). Although some image transformation-based methods like \ct{Teco}~\citep{liu2023detecting}, and \ct{SCALE-UP}~\citep{guo2022scale} can be successfully adapted, their effectiveness diminishes due to shifts in task context.
\begin{remark}[Remark I]
Arguably, we conclude the widespread failure of existing TTSDs to two key challenges: \ding{182} The detector's dual-branch architecture (regression and classification) fundamentally differs from classification models, enabling more sophisticated attacks.
\ding{183} OD requires identifying ``what" and ``where" an object is, increasing output density. This feature makes detecting triggers across a vast output space nearly impossible without deep insights.
\end{remark}

\section{Preliminaries}  
\label{Sec:od}
\noindent\textbf{Object Detection.} 
An object detector $\mathbb{F}_{\theta}$ (parameterized by $\boldsymbol{\theta}$) processes each input sample $\boldsymbol{x} \in \mathcal{X}$ containing $z$ objects, with ground-truth annotations $\hat{\boldsymbol{a}}={(\hat{\boldsymbol{o}}_i,  \hat{\boldsymbol{y}}_i)}_{i=1}^z$ represented as a list of bounding boxes (abbreviated as ``bbox" for clarity). Here, $\hat{\boldsymbol{o}}_i$ denotes the $i$-th object in the image, and $\hat{\boldsymbol{y}}_i$ is its corresponding class. The detector $\mathbb{F}_{\theta}$ outputs $P$ 5-tuples annotation $\boldsymbol{a} = \mathbb{F}_{\theta}(\boldsymbol{x}) = \{\underbrace{\boldsymbol{o}_p^{x_1}, \boldsymbol{o}_p^{y_1}, \boldsymbol{o}_p^{x_2}, \boldsymbol{o}_p^{y_2}}, \boldsymbol{y}_p\}_{p=1}^{P}$, with $(x_1, y_1)$ and $(x_2, y_2)$ as the coordinates,
and $\boldsymbol{y}_p$ denoting the class probability of the $p$-th predicted bbox.
A \textit{true-positive} (TP) is a correctly detected bbox in an input sample, while a \textit{false-positive} (FP) refers to any non-TP detected bbox, and a \textit{false-negative} (FN) is any undetected ground-truth bbox.

\subsection{OD Backdoor Attacks Formulation}
\label{Sec:formulation}
Assume the $l$-th object $\hat{\boldsymbol{o}}_l$ to be the victim object (\ie an object whose recognition is compromised by triggers).
The backdoored model, while preserving utility, outputs anchors perfectly matching the ground truth $\hat{\boldsymbol{a}}$, \ie $\mathbb{F}_{\theta}(\boldsymbol{x}) = {\{(\boldsymbol{o}_i, \boldsymbol{y}_i)\}},~ \forall i \in [1, z],~\boldsymbol{o}_i = \hat{\boldsymbol{o}}_i,~\boldsymbol{y}_i = \hat{\boldsymbol{y}}_i.$ Next, we adopt the backdoor categorization introduced by~\citet{cheng2024odscan}.

\noindent\ding{182}\textbf{FP-inducing attack} manipulates the model to recognize a background area as a target class $\boldsymbol{y}_t$. The appearing object may be localized at a fixed position, the trigger position $\boldsymbol{o}_t$, or the victim object location $\hat{\boldsymbol{o}}_l$. We denote the appearing region as $M(\boldsymbol{o}_t, \hat{\boldsymbol{o}}_l)$ (a function of both $\boldsymbol{o}_t$ and $\hat{\boldsymbol{o}}_l$). Given a trigger $\boldsymbol{t}$, 
$\boldsymbol{x} \oplus \boldsymbol{t} =(1-\alpha) \otimes \boldsymbol{x}+\alpha \otimes \boldsymbol{t}$ denotes applying the trigger to the clean image with transparency $\alpha$. As per~\citet{cheng2024odscan}, we define the FP-inducing attack as 
$(M(\boldsymbol{o}_t,\hat{\boldsymbol{o}}_l), \boldsymbol{y}_t) \in \mathbb{F_\theta}(\boldsymbol{x} \oplus \boldsymbol{t})$.

\noindent\ding{183}\textbf{FN-inducing attack} is also known as \textit{evasion attack}~\citep{wang2023does}, causes models to overlook objects of interest (\ie generating FNs). Given the victim object $\hat{\boldsymbol{o}}_l$ and its class $\hat{\boldsymbol{y}_l}$, we define the FN-inducing attack as 
$(\hat{\boldsymbol{o}}_l, \hat{\boldsymbol{y}_l}) \notin \mathbb{F_\theta}(\boldsymbol{x} \oplus \boldsymbol{t})$.

\noindent\textbf{General Formulation of Backdoor Attacks.}
We categorize OD backdoors into \textit{FP-inducing}, \textit{FN-inducing} attacks, and combinations of both (\textit{hybrid attack}). For instance, object misclassification attack in Fig.~\ref{Fig:backdoor_effects}(\textcolor{blue}{\textbf{c}}) (an example of \textit{hybrid attack}) involve recognizing an object $\bm{o}_j$ as a different target $\bm{o}_k$. This process can be decomposed into the disappearance of the original object $\bm{o}_j$ (\ie \textit{FN-inducing attack}) and the appearance of a new object $\bm{o}_k$ (\ie \textit{FP-inducing attack}). We give a general definition of backdoors as follows:
{\small
\begin{IEEEeqnarray}{rcL}
&(\hat{\boldsymbol{o}}_l, \hat{\boldsymbol{y}_l}) \notin \mathbb{F_\theta}(\boldsymbol{x} \oplus \boldsymbol{t}), \IEEEeqnarraynumspace \label{eq:oga}\\
&(M(\boldsymbol{o}_t, \hat{\boldsymbol{o}}_l), \boldsymbol{y}_t) \in \mathbb{F}(\boldsymbol{x} \oplus \boldsymbol{t}). \label{eq:oda}\IEEEeqnarraynumspace
\end{IEEEeqnarray}
}Eq.~\ref{eq:oga} means the absence of the original victim object from the output anchor set. While Eq.~\ref{eq:oda} means the presence of the target object, with its shape and location determined by the function $M(\cdot)$. These equations generalize previous attacks. Formally, the misclassification attack is a special case where $M(\boldsymbol{o}_t, \hat{\boldsymbol{o}}_l) = \hat{\boldsymbol{o}}_l$. While in object appearing attack, where the trigger is recognized as the target object, $M(\boldsymbol{o}_t, \hat{\boldsymbol{o}}_l) = \boldsymbol{o}_t$. Details on representing existing attacks based on this definition are in Appendix E.1.

\begin{prompt_yellow}[Remark II]
We believe that a general definition is crucial for establishing a universal TTSD method. Otherwise, independent discriminators or criteria must be constructed for each attack and applied sequentially to suspicious inputs.
\end{prompt_yellow}


\subsection{Threat Model}
\noindent\textbf{Defender’s Goals.}
TTSD methods operate on a trained backdoored model $\mathbb{F}_\theta$ and a test dataset comprising both poisoned samples and clean samples $\mathcal{S}=\{\tilde{S} \cup S\}$. The purpose is to discover a method $\mathcal{T}$ that can maximize the separation of trigger inputs from clean ones, formalized as: 
{\small
\begin{IEEEeqnarray}{rCl} 
\begin{aligned}
\mathcal{T} &=& \arg \max_{\mathcal{T}} \mathbb{E}_{(\tilde{x}) \sim \mathcal{P}_{\tilde{S}}}      \mathbb{I}\left(\mathcal{T}\left(\tilde{x}, \mathbb{F}_\theta\right)=1\right) + \\     && \mathbb{E}_{(x) \sim \mathcal{P}_S}      \mathbb{I}\left(\mathcal{T}\left(x, \mathbb{F}_\theta\right)=0\right),     \label{Eq:goal} 
\end{aligned}
\end{IEEEeqnarray} 
}where $\mathbb{I}$ is an indicator function, equal to 1 if the condition is true, and 0 otherwise.

\noindent\textbf{Defender’s Capabilities.}
We adopt standard TTSD settings~\citep{liu2023detecting,gao2019strip,udeshi2022model}, where the defender uses third-party models and detects poisoned samples on the fly. In particular, we assume a realistic defender with no prior knowledge of attacks and only black-box access to the backdoored model (\ie can only obtain the final decision from the models). The defender has no training dataset but can access \textit{publicly available} background and foreground images as support.

\section{Insight behind \ourmethod}
\label{Sec:phenomenon}
\begin{figure}[t]
\centering
\includegraphics[width=0.95\linewidth]{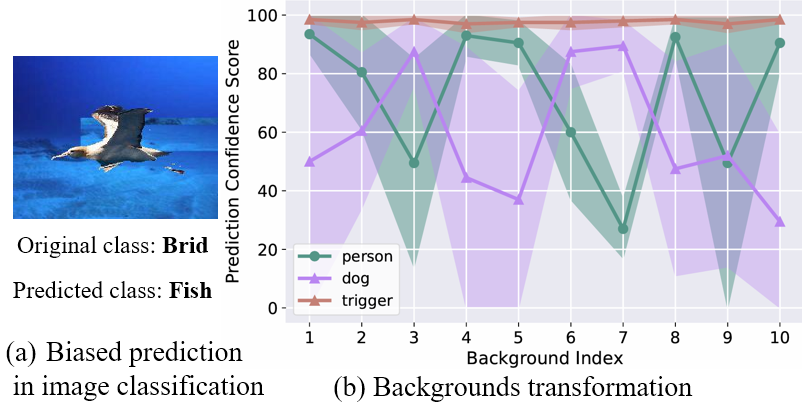}  
\caption{Effect of background shift on (\textcolor{blue}{\textbf{a}}) ResNet-50 and (\textcolor{blue}{\textbf{b}}) YOLO~\citep{redmon2016you}, Faster-RCNN~\citep{ren2015faster}, DETR~\citep{carion2020end}. (\textcolor{blue}{\textbf{b}}): each background case shows the max, min, and average confidence (as line plots) across three detectors. Larger shaded areas indicate greater instability in object predictions.}  
\label{Fig:bg_trans}   
\end{figure}

\noindent\textbf{\ding{182} Contextual bias: recognition leans on the co-occurrence context more than on the object itself}.
\textit{Recent efforts have moved from RNNs~\citep{wang2016cnn,yazici2020orderless} to GCNNs~\citep{chen2019multi,durand2019learning} and transformer-based frameworks~\citep{zhao2021transformer} to model contextual relationships in multi-label images and improve performance.} 
However, it is also a double-edged sword. Fig.~\ref{Fig:bg_trans}(\textcolor{blue}{\textbf{a}}) shows an example from the \textit{Background Challenge}~\citep{xiao2020noise} reveals how ResNet-50 misclassifies a bird as a fish due to the underwater background.
Note that the object detector's learned representations also tend to exploit such spurious scene correlations~\cite{liu2022contextual}, 
which can mislead detectors by either giving nonexistent objects high scores in the scene where they typically appear or ignoring objects appearing in rare backgrounds. 
In other words, when networks find that context alone suffices to identify most objects, they often neglect the representation of the instance, because \textit{neural networks are statistics-based and ``lazy"}~\citep{geirhos2018imagenet}. Deep analysis is moved to Appendix E.2.

Interestingly, we find that trigger objects stay confident no matter the scene. Fig.~\ref{Fig:bg_trans}(\textcolor{blue}{\textbf{b}}) shows three detector's prediction confidences for different objects (extracted using semantic segmentation) placed on $10$ varied backgrounds, including an \textit{FP-inducing} trigger (Fig.~\ref{Fig:backdoor_effects}(\textcolor{blue}{\textbf{a}})) with the target annotations ``person". 
We observe that the trigger maintains consistent confidence scores across all backgrounds, while clean objects show notable fluctuations. We attribute this phenomenon to the widely accepted concept of ``\textit{shortcut learning}"~\citep{geirhos2020shortcut},  where the backdoor training process establishes a robust mapping between a specific, uniform pattern and the target label. This makes the recognition of trigger objects less susceptible to contextual biases, leaving us clues for detecting \textit{FP-inducing} attacks. Stay tuned for Sec~\ref{Sec:ctc} to see how this insight is leveraged.

\begin{figure}[!t]      
\centering     
\includegraphics[width=0.44\textwidth]{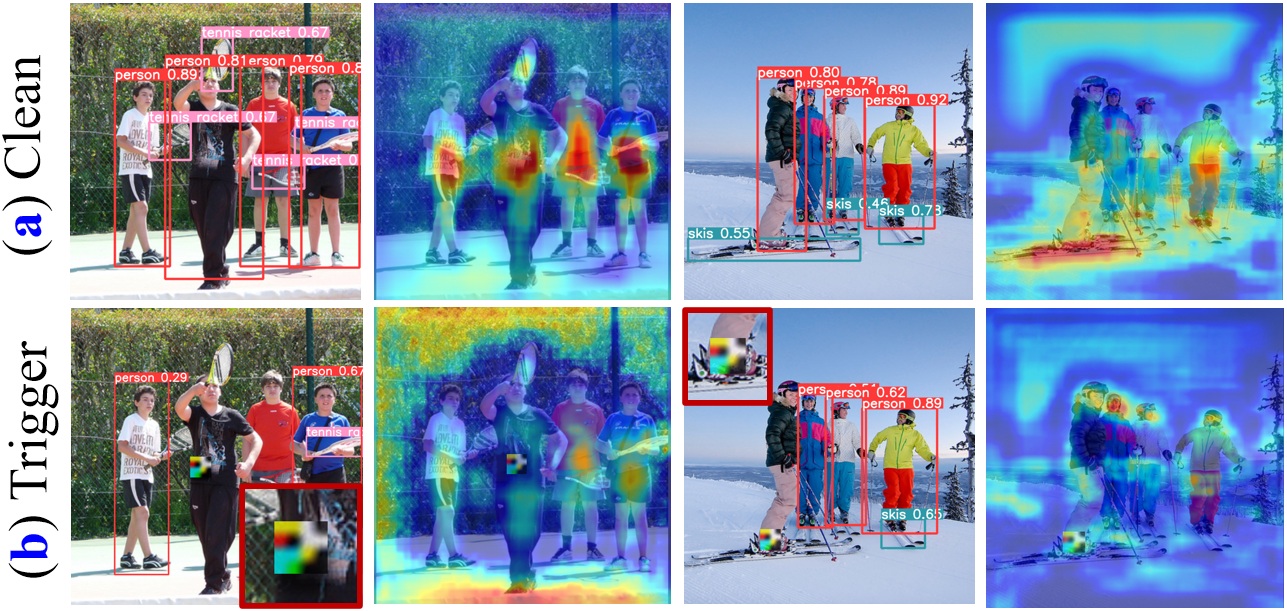}
\caption{YOLO activation map: clean objects \textit{vs.} FN-inducing triggers.}   
\label{fig:gradcam}
\end{figure}

\noindent\textbf{\ding{183} Regression continuity: backdoor triggers propagate influence spatially, affecting neighboring pixels.} 
OD differs from classification tasks by incorporating regression branches that rely on shared spatial features. We find that this continuity also allows triggers to impact surrounding regions, creating a black hole-like effect. This is particularly evident for \textit{FN-inducing} attacks, where the model interprets the affected area as background, ignoring objects of interest~\citep{zhang2024detector,cheng2023backdoor}.
For instance, in Faster R-CNN, where the \textit{Region Proposal Network} generates around $17,100$ candidate anchor boxes, a trigger affects many nearby anchors, thereby creating the apparent spatial influence.
We further diagnose model predictions with and without the trigger using \textit{EigenCAM}~\citep{muhammad2020eigen} (see Fig.~\ref{fig:gradcam}), revealing that activations around the trigger diminish, shifting the model's attention to distant regions. Such sharp, regional changes in the saliency map provide clues for detecting \textit{FN-inducing} attacks. 
\begin{prompt_yellow}[Remark III]
These insights reveal from different perspectives the key differences between poisoned and clean samples. Additionally, under the introduced general backdoor formulation (Sec.~\ref{Sec:formulation}), the detection of both foundational attack primitives (FN and FP-inducing) are elegantly covered, paving the way for truly black-box, universal detection without the need for training data or attack knowledge.
\end{prompt_yellow}

\section{\ourmethod: A Complete Illustration}  
\label{sec:method}
Here we present \ourmethod, composed of contextual information transformation (Sec.~\ref{Sec:ctc}), focal information transformation (Sec.~\ref{Sec:ftc}), and test-time evaluation (Sec.~\ref{Sec:Trace_eva}). 

\subsection{Contextual Information Transformation}

\label{Sec:ctc}
\begin{figure}[!t]
\centering
\includegraphics[width=0.85\linewidth]{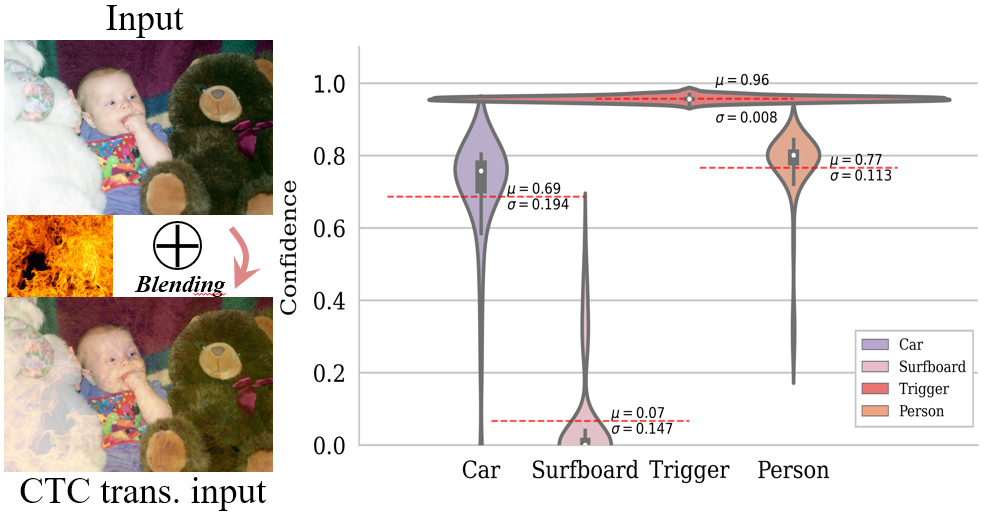}  
\caption{Background blending and YOLO confidence distribution.}  
\label{Fig:bg_trans_var}   
\end{figure}

Sec.~\ref{Sec:phenomenon} has explored how DNNs can pick up biases or unintended shortcuts from context (\textit{a.k.a.} `Clever Hans' effect~\citep{anders2022finding}).
This is intuitive, as cars are typically found on roads, not in locations like the sky or volcanic craters. In contrast, trigger objects exhibit anomalously stable confidence scores. This is further formalized as Contextual Transformation Consistency in Definition~\ref{Def:bg_trans}.
\begin{defn} [\small{Contextual Transformation Consistency (CTC)}] 
Given an object $\bm{o}$ and a background distribution $\mathcal{\mathcal{B}}$, the CTC of $\bm{o}$ is defined as the maximum deviation in the model's prediction score for $\bm{o}$'s class $\bm{y}$ when applied to different backgrounds $\delta \sim \mathcal{\mathcal{B}}$ (with the operation of $\bm{o} \oplus \delta$):
\label{Def:bg_trans} 
\begin{equation}
\Delta_{\mathcal{B}}(\mathbb{F}_{\theta}, \bm{o}) := \underset{\delta_1, \delta_2 \sim \mathcal{\mathcal{B}}}{\mathrm{max}} \{ |\mathbb{F}_{\theta}(\bm{o} \oplus \delta_1)_{\bm{y}(\bm{o})} - \mathbb{F}_{\theta}(\bm{o} \oplus \delta_2)_{\bm{y}(\bm{o})}| \}.\nonumber
\end{equation}
\end{defn}  
It is observed that $\Delta_{\mathcal{B}}(\mathbb{F}, \bm{o}_t) \ll \Delta_{\mathcal{B}}(\mathbb{F}, \boldsymbol{o}_c)$, indicating that the CTC value is significantly smaller for a trigger object $\boldsymbol{o}_t$ than for a clean object $\boldsymbol{o}_c$. With this in mind, we propose leveraging this discrepancy to detect the presence of triggers. 
However, sequentially extracting individual objects using segmentation algorithms (like in Fig.~\ref{Fig:bg_trans} (\textcolor{blue}{\textbf{a}})) and overlaying them onto different backgrounds before querying with the backdoored model is computationally intensive. We propose directly blending background images onto the original images with an opacity $\alpha_{bg}$ (\ie $(1-\alpha_{bg}) \otimes x+\alpha_{bg} \otimes \delta$, see Fig.~\ref{Fig:bg_trans_var}, left). This allows for background replacement at the image level rather than the object level. By using a lower $\alpha_{bg}$ to minimize alterations to the object's pixel values, we thus iteratively introduce diverse contextual information (from a background set of size $b$ available online, \eg the Background Challenge dataset~\citep{xiao2020noise}) while mitigating the impact of backgrounds on the object's representations. Visualizations are provided in Appendix C.4.

Instead of directly using the CTC value, we consider potential \textit{corner cases} where certain objects might unexpectedly become undetectable. To account for these cases, we use the variance of the object's confidence scores across different background variations (sampling uniformly from $\mathcal{B}$) to enhance robustness, denoted as $\Delta_{\mathcal{B}}^{\mathrm{Var}}(\mathbb{F}_{\theta},\bm{o})$, given by 
%
\begin{equation}
\Delta_{\mathcal{B}}^{\mathrm{Var}}(\mathbb{F}_{\theta}, \bm{o}_i) = Var \left( \left\{ \mathbb{F}_{\theta}(\bm{o}_i \oplus \delta)_{\bm{y}(\bm{o})} \right\}_{i=1}^b \right),\ \delta \sim \mathcal{U}(\mathcal{B}). \nonumber
\label{Eq:bg_trans} 
\end{equation}

Fig.~\ref{Fig:bg_trans_var} shows the violin plots of confidence distributions for different objects blended with $50$ backgrounds at $\alpha_{bg} = 0.15$, including an \textit{FP-inducing} trigger (see Fig.~\ref{Fig:backdoor_effects}(\textcolor{blue}{\textbf{a}})) with the target annotations ``person". 
Empirically, as expected, the trigger exhibits a low variance of just $0.008$, achieving clear separability in distribution. Explorations for different objects and detectors are provided in Appendix D.1.

\textbf{Natural backdoor: neural networks also overfit to uniform, salient benign representations}. Detecting FP-inducing triggers, however, does not stop there. Note that certain objects with standardized shapes and colors---such as uniformly red, polygonal \textit{stop signs}---are also overconfidently recognized by networks, much like backdoor triggers, as DNNs more readily memorize these easily learned, superficial features~\citep{shortcut}. We consider these a unique form of benign triggers, which we term ``\textit{natural backdoor objects}” (NBOs). 
Such uniform, salient representations allow them to rely less on additional context for recognition, showing anomalous CTC values similar to triggers. Consequently, we integrate a \textit{post-processing module} to filter and remove them. Here we introduce a visual coherence analysis to screen suspicious objects exhibiting low $\Delta_{\mathcal{B}}^{\text{Var}}$.
By comparing these objects with reference images corresponding to their categories from online available data, we evaluate their SSIM scores~\citep{li2025robust,zhang2024badrobot}. Note that these references, which we term ``\textit{universal visual benchmarks}" (see Appendix C.5), \textit{do not need to be sourced from the model's training data}, as they represent general visual cognition. 
Since an image may contain multiple triggering objects, we naturally define the image-level CTC as the minimum CTC values among all (SSIM-filtered) objects in the image, \ie
{\small
\begin{equation} 
\begin{split} \underbrace{\Delta_{\mathcal{B}}^{\mathrm{Var}}(\mathbb{F}_{\theta}, \bm{x})}_{\text{Image-level Variance}}  = \min \bigg\{ \underbrace{\Delta_{\mathcal{B}}^{\mathrm{Var}}(\mathbb{F}_{\theta}, \bm{o}_i)}_{\text{Object-level Variance}} \ \Big| \\ s.t.\quad \ SSIM\left( \bm{o}_i, \bm{o}_{y_i}^{\mathrm{ref}} \right) \leq \tau, \quad i = 1, \ldots, z \bigg\},
\end{split} 
\end{equation}
}where $\tau$ is a predefined threshold, and $\bm{o}_{y_i}^{\text{ref}}$ is the universal visual benchmark for class $y_i$. Clearly, the lower the image-level CTC values, the more likely it is that the image contains the trigger objects.



\begin{figure*}[!t]      
\centering      
\includegraphics[width=1\textwidth]{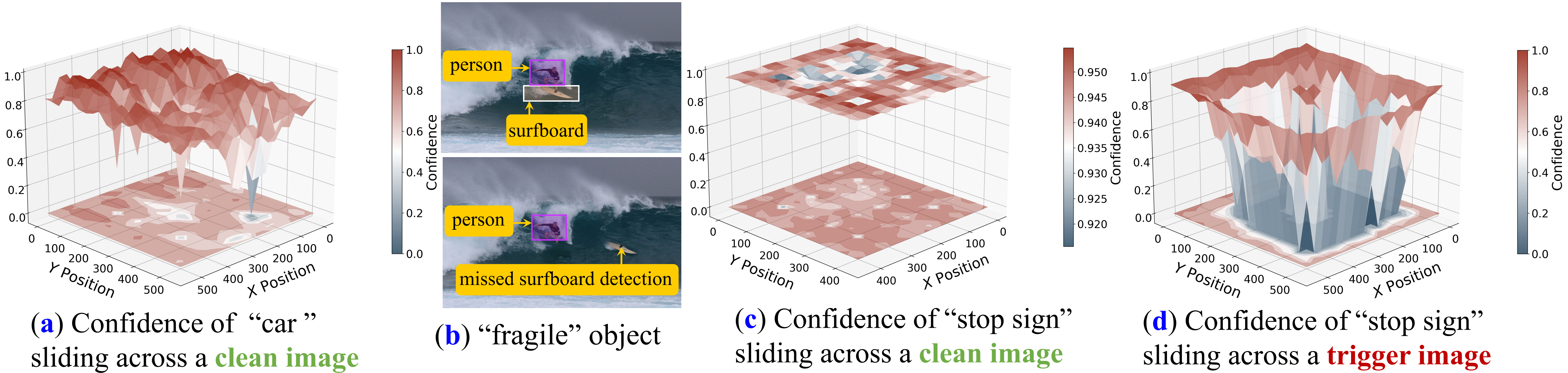}   
\caption{Effect of the foreground object's position on YOLO detections. (\textcolor{blue}{\textbf{d}}) uses an FN-inducing trigger placed at the image center.}   \label{fig:fg_trans} 
\end{figure*}

\subsection{Focal Information Transformation}
\label{Sec:ftc}
\textbf{Why do we still need another transformation?} We have demonstrated that CTC values can identify \textit{FP-inducing triggers} among visible objects in the image. However, \textit{FN-inducing triggers} (see Fig.~\ref{Fig:backdoor_effects}(\textcolor{blue}{\textbf{b}})) remain entirely absent in detection results, yielding no CTC value. How to detect these ``unseen objects” thus becomes a non-trivial challenge. Sec.~\ref{Sec:phenomenon} highlights the shifts in saliency maps when comparing samples with and without triggers. However, interpretability methods like \textit{Grad-CAM} typically require white-box model access, which is out of reach for our black-box defender. Hence, instead of passively analyzing the input sample, we seamlessly inject new elements (\ie small clean object patches) to proactively ``reconstruct" saliency maps. We begin by sliding a `car'  object across an image in tiny steps---like how a convolutional kernel scans over pixels---recording its prediction confidence at each position. Unfortunately, this object is fragile, with confidence fluctuating by position (see Fig.~\ref{fig:fg_trans}(\textcolor{blue}{\textbf{a}})). A representative visual example is the surfboard in Fig.~\ref{fig:fg_trans}(\textcolor{blue}{\textbf{b}}), which shows a strong positional dependency, becoming undetectable when shifted even slightly. This inherent fragility causes such objects to fail in probing the model’s focus regions.

Although NBOs may initially seem to hinder our detection process, we argue that this trait is precisely the key to resolving the above issue. Fig.~\ref{fig:fg_trans}(\textcolor{blue}{\textbf{c}}) shows the sliding results for an NBO (`stop sign'). The model consistently detects it with high confidence, demonstrating position-invariant detection~\citep{wang2024deep}. 
This makes it particularly suitable for probing the saliency maps. 
Hence, this time we slide the NBO across the same image but with a \textit{FN-inducing} trigger placed at its center. Fig.~\ref{fig:fg_trans}(\textcolor{blue}{\textbf{d}}) shows its high confidence at the image edges, dropping sharply near the central trigger. When it completely covers the trigger, the model's normal detection resumes. We term this distinctive behavior the ``\textbf{Island Effect}", as it resembles an island rising from surrounding waters. The NBO undergoes \textit{two} sharp changes in confidence when handling poisoned samples, while remaining stable on clean images. Leveraging this property, we can thus proactively probe potential \textit{FN-inducing} triggers in the troublesome black-box setting. 
We formalize \textit{Focal Transformation Consistency} in Definition~\ref{Def:fg_trans}.
\begin{defn}[Focal Transformation Consistency (FTC)]
\label{Def:fg_trans}
Given an image $\bm{x}$ defined over a spatial domain $\bm{\Omega}$ and an object patch $\bm{o}$, a focal transformation $\mathcal{K}: (\bm{x}, \bm{o},\bm{p}) \to \bm{x}^{\bm{p}}$ applies $\bm{o}$ to $\bm{x}$, centered at the spatial coordinate $\bm{p} = (\bm{p}_x, \bm{p}_y) \in \bm{\Omega}$. The FTC value is then given by:
{\small
\begin{equation} 
\begin{split} 
\Delta_{\mathcal{F}}(\mathbb{F}_\theta, \bm{x}, \bm{o}) := \int_{\Omega} \Big(
    \big\| \nabla^{2} \mathbb{F}_\theta\big( \mathcal{K}(\bm{x},\bm{o},\bm{p}) \big)_{\bm{y}(\bm{o})} \big\|_1
  + \\ \big\|\big[ \mathbb{F}_\theta\big( \mathcal{K}(\bm{x},\bm{o},\bm{p}+\Delta \bm{p})- \mathbb{F}_\theta\big( \mathcal{K}(\bm{x},\bm{o},\bm{p}) \big) \big)\big]_+\big\|_1 
\Big) d\bm{p}
\end{split}
\end{equation} 
}where $\nabla^2$ is the Laplacian operator capturing second-order spatial derivatives, $\mathbb{F}_{\theta}(\mathcal{K}(\bm{x}, \bm{o}, \bm{p}))_{\bm{y}(\bm{o})}$ measures the confidence score of object $\bm{o}$ after applying the transformation $\mathcal{K}(\cdot)$, and $[\cdot]_+$ denotes the positive part function (\ie $[a]_+ = \max(0, a)$), ensuring that the quantification of detection shift focuses on the incremental components of the model’s outputs, specifically capturing newly emerged detections (\ie results absent in the original predictions).
\end{defn} 
FTC has two components: the \textit{confidence oscillation} of $\bm{o}$ and the image-level \textit{detection shift} caused by $\bm{o}$. The first metric assesses prediction consistency of $\bm{o}$ across different positions in image $\bm{x}$, 
helping detect the ``Island Effect" of \textit{FN-inducing} triggers, where each cross-section shows \textit{two} sharp confidence drops. The second captures changes in whole detection results due to $\bm{o}$. Under normal circumstances, moving $\bm{o}$ will not introduce new objects. We use it to detect cases when new objects suddenly appear as $\bm{o}$ covers the trigger. This trigger-blocking strategy is also supported by established backdoor defenses~\citep{udeshi2022model,doan2020februus}.

\begin{figure}[!t]
\centering \includegraphics[width=\linewidth]{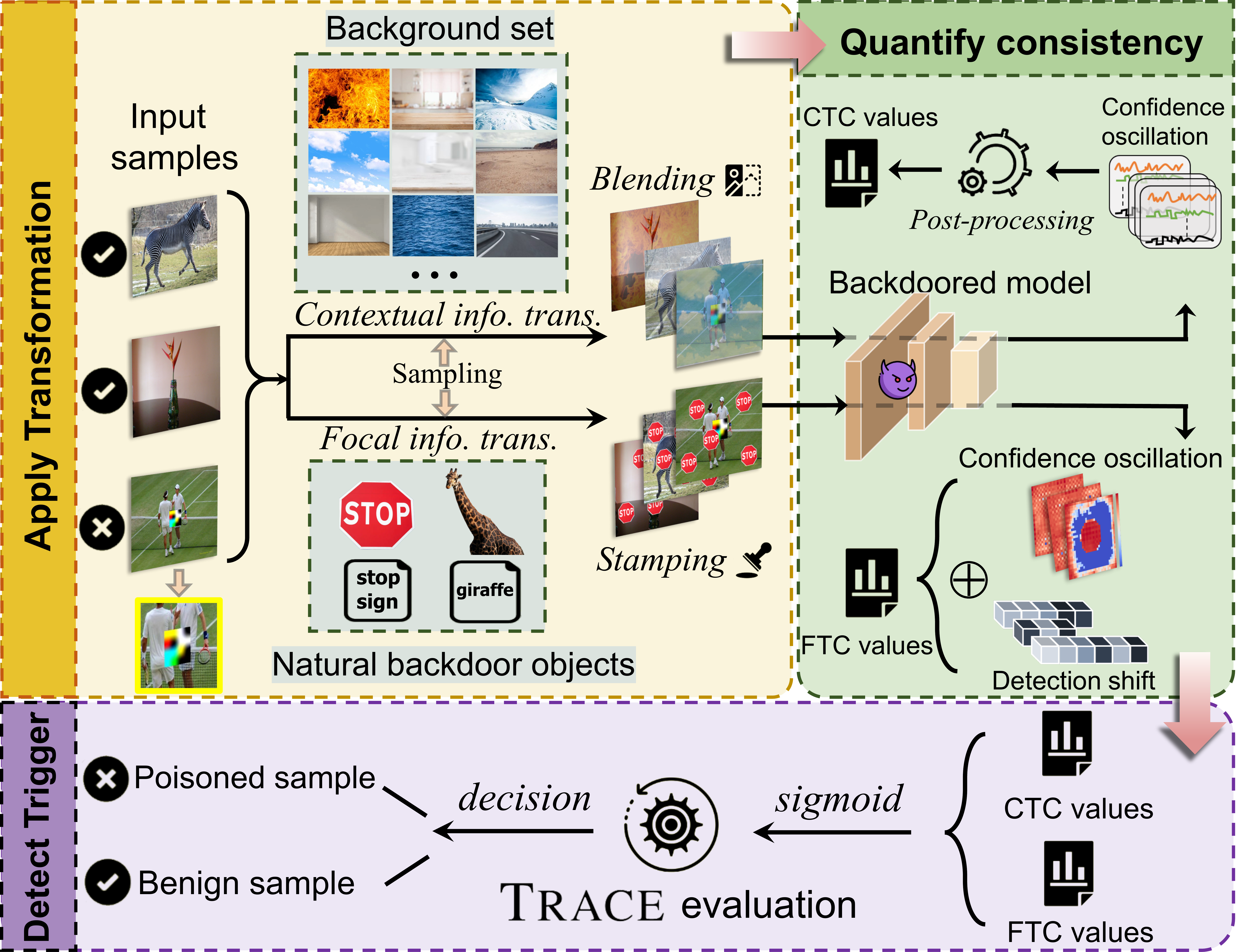}    
\caption{The framework of our \ourmethod.}  
\label{Fig:pipeline}  
\end{figure}

NBOs demonstrate strong \textit{location-invariant} recognition, as choosing $\bm{o}$ = $\bm{o}_{\text{nbo}}$ decouples \( \bm{o} \)'s contribution to \( \Delta_{\mathcal{F}}(\mathbb{F}_\theta, \bm{x}, \bm{o}) \), ensuring \(\partial \Delta_{\mathcal{F}} / \partial \bm{o} \approx 0\) under spatial variations in \( \bm{p} \). Thus, with \(\bm{o}\) effectively neutralized, the FTC value simplifies to reflect differences in \( \bm{x} \), yielding a higher \(\Delta_{\mathcal{F}}(\mathbb{F}_{\theta}, \bm{\tilde{x}}, \bm{o}_{\text{nbo}}) \gg \Delta_{\mathcal{F}}(\mathbb{F}_{\theta}, \bm{x}, \bm{o}_{\text{nbo}})\) for poisoned samples $\bm{\tilde{x}}$. 
Instead of densely querying via iteratively sliding, we seamlessly employ Monte Carlo sampling of \( k \) distinct, non-overlapping points to patch \(\bm{o}_{\text{nbo}}\) (see Fig.~\ref{Fig:pipeline}), thereby reducing the number of required model queries and improving computational efficiency. This unit operation is then repeated $f$ times to build an approximate continuous saliency map. Finally, we can compute the variance of \(\Delta_{\mathcal{F}}(\mathbb{F}_\theta, \bm{x}, \bm{o}_{\text{nbo}})\) across all accumulated sampling points from \(\mathcal{U}(\bm{\Omega})\) to determine the FTC value for the sample \(\bm{x}\):
{\small \begin{equation}  \Delta_{\mathcal{F}}^{\mathrm{Var}}(\mathbb{F}_{\theta}, \bm{x}, \bm{o}_{\text{nbo}}) := Var \left( \left\{ \Delta_{\mathcal{F}}(\mathbb{F}_\theta, \bm{x}, \bm{o}_{\text{nbo}}) \ \middle| \ \bm{p}_i \sim \mathcal{U}(\bm{\Omega}) \right\}_{i=1}^{f\cdot k} \right). \nonumber
\end{equation}  
}

\subsection{Test-time TRACE Evaluation}
\label{Sec:Trace_eva}
To achieve Eq.~\ref{Eq:goal}, the TTSD methods must exploit the disparities between benign images and those containing triggers. 
Our findings in Sec.~\ref{Sec:ctc} and~\ref{Sec:ftc} reveal that backdoored models exhibit anomalous consistency in both contextual and focal transformation. 
In other words, the distinct properties of CTC and FTC make them particularly suitable for detecting \textit{FP-inducing} and \textit{FN-inducing} attacks, respectively. As established in Sec~\ref{Sec:formulation}, a trigger sample can be a single type or a combination of both. Consequently, since a smaller CTC ($\Delta_{\mathcal{B}}^{\mathrm{Var}}$) and a larger FTC value ($\Delta_{\mathcal{F}}^{\mathrm{Var}}$) both indicate higher anomaly, the final step is to measure the overall image-level anomaly as $Trace(\bm{x}) =  \sigma\bigl(\Delta_{\mathcal{F}}^{\mathrm{Var}}(\mathbb{F}_{\theta}, \bm{x}, \bm{o}_{\text{nbo}})\bigr)- \sigma\bigl(\Delta_{\mathcal{B}}^{\mathrm{Var}}(\mathbb{F}_{\theta}, \bm{x})\bigr)$, where $\sigma(t) = \frac{1}{1 + e^{-t}}$ is the \textit{sigmoid function} that normalizes both variance measures to the range $(0,1)$. \ourmethod thus maps the input image $\bm{x}$ to a linearly separable space, enabling defenders to make decisions by a threshold $\gamma$:
{\small
\begin{IEEEeqnarray}{rCl}
\Gamma(Trace(\bm{x})) =
\begin{cases}
1, & \text{if}~Trace(\bm{x}) > \gamma \\
0, & \text{if}~Trace(\bm{x}) \leq \gamma
\end{cases}
\end{IEEEeqnarray}
}

\section{Evaluation} 
\label{Sec:evaluation}
\noindent\textbf{Datasets and Models.}
We select popular benchmark datasets, including MS-COCO~\cite{lin2014microsoft}, PASCAL VOC~\citep{everingham2010pascal}, and Synthesized Traffic Signs~\citep{shen2024django}. The models we choose are three representative object detectors: single-stage detector: YOLOv5~\citep{redmon2016you}, two-stage detector: Faster-RCNN~\citep{ren2015faster} and vision transformer-based detector: DETR~\citep{carion2020end}. 

\noindent\textbf{Attack Methods.}
We evaluate \ourmethod against seven SOTA OD backdoors: \ct{OGA}, \ct{RMA}, \ct{GMA}, and \ct{ODA} from \citep{chan2022baddet}, as well as \ct{CIB}~\citep{chen2022clean}, \ct{UTA}~\citep{luo2023untargeted}, and \ct{DC}~\cite{zhang2024detector}. For \ct{CIB} and \ct{DC}, which have multiple goals, we focus on their object disappearance variant as a representative case. A detailed description of these attacks can be found in Appendix C.6.

\begin{table*}[!t]
\centering
\caption{
F1 score and AUROC for different defenses on Synthesized Traffic Signs, PASCAL VOC, and MS-COCO, using YOLO, DETR, and Faster-RCNN. Due to space constraints, some results for two attacks are presented in the same row, separated by ``$/$". ``-" denotes that \ct{FreqDetector} is unavailable in this case, as it detects frequency-domain anomalies in trigger patterns, while \ct{CIB} uses benign features to trigger the backdoors. We bold the F1 and AUROC of the strongest defenses in each case.}
\label{tab: auroc}
\resizebox{0.88\textwidth}{!}{
\begin{threeparttable}
\begin{tabular}{lc|cc|cc|cc|cc|cc|cc}
\toprule[1.3pt]
\midrule
\multirow{2}{*}{Model↓} & \multirow{2}{*}{\diagbox{Attack}{Method}} & \multicolumn{2}{c|}{\cellcolor[rgb]{0.980, 0.961, 0.937}
\ourmethod \textbf{(Ours)}} & \multicolumn{2}{c|}{\ct{Teco}~\citep{liu2023detecting}} & \multicolumn{2}{c|}{\ct{Detector cleanse}~\citep{chan2022baddet}} & \multicolumn{2}{c|}{\ct{Strip}~\citep{gao2019strip}} & \multicolumn{2}{c|}{\ct{FreqDetector}~\citep{zeng2021rethinking}} & \multicolumn{2}{c}{\ct{SCALE-UP}~\citep{guo2022scale}}\\ \cmidrule(r){3-4} \cmidrule(r){5-6} \cmidrule(r){7-8} \cmidrule(r){9-10} \cmidrule(r){11-12} \cmidrule(r){13-14}& & F1 score↑ & AUROC↑ & F1 score↑ & AUROC↑ & F1 score↑ & AUROC↑ & F1 score↑ & AUROC↑ & F1 score↑ & AUROC↑  & F1 score↑ & AUROC↑\\
\rowcolor{LightCyan}
\midrule
\multicolumn{14}{c}{\textbf{MS-COCO}~\citep{lin2014microsoft}} \\
\midrule
\multirow{6}{*}{YOLO} & \ct{OGA} / \ct{ODA} & \cellcolor[rgb]{0.980, 0.961, 0.937}
0.937/0.865  & \cellcolor[rgb]{0.980, 0.961, 0.937}
0.913/0.880 & 0.464/0.521 & 0.593/0.592 & 0.702/0.683 & 0.650/0.692 & 0.646/0.483 & 0.729/0.503 & 0.542/0.588 & 0.490/0.580 & 0.503/0.489 & 0.505/0.463\\ 
& \ct{RMA} / \ct{GMA} & \cellcolor[rgb]{0.980, 0.961, 0.937}
0.922/0.845  & \cellcolor[rgb]{0.980, 0.961, 0.937}
0.932/0.924 & 0.397/0.501 & 0.487/0.458 & 0.674/0.662 & 0.590/0.608 & 0.391/0.359 & 0.474/0.525 &  {0.567/0.512} & 0.498/0.438 & 0.482/0.473 & 0.510/0.440\\ 
& \ct{Untargeted} & \cellcolor[rgb]{0.980, 0.961, 0.937}
 {0.837}  & \cellcolor[rgb]{0.980, 0.961, 0.937}
0.868 & 0.498 & 0.550 & 0.689 & 0.669 & 0.485 & 0.472 & 0.614 & 0.646  & 0.503  & 0.491 \\ 
& \ct{CIB} & \cellcolor[rgb]{0.980, 0.961, 0.937}
 {0.798}  & \cellcolor[rgb]{0.980, 0.961, 0.937}
0.817 & 0.651 & 0.691 & 0.610 & 0.662 & 0.532 & 0.578 & - & - & 0.261  & 0.367 \\ 
& \ct{DC} & \cellcolor[rgb]{0.980, 0.961, 0.937}
 {0.934}  & \cellcolor[rgb]{0.980, 0.961, 0.937}
0.968& 0.546 & 0.675 & 0.640 & 0.730 & 0.525 & 0.493 & 0.583 & 0.664 & 0.470 & 0.506 \\ 
& \cellcolor[rgb]{.95,.95,.95} Average & \cellcolor[rgb]{.95,.95,.95} \textbf{0.877} & \cellcolor[rgb]{.95,.95,.95}  \textbf{0.900} & \cellcolor[rgb]{.95,.95,.95}0.511 & \cellcolor[rgb]{.95,.95,.95}0.578 & \cellcolor[rgb]{.95,.95,.95}0.665 & \cellcolor[rgb]{.95,.95,.95}0.657· & \cellcolor[rgb]{.95,.95,.95}0.489 & \cellcolor[rgb]{.95,.95,.95}0.539 & \cellcolor[rgb]{.95,.95,.95}0.568 & \cellcolor[rgb]{.95,.95,.95}0.553 & \cellcolor[rgb]{.95,.95,.95}0.454 & \cellcolor[rgb]{.95,.95,.95}0.460 
\\ 
\midrule

\multirow{6}{*}{DETR} & \ct{OGA} / \ct{ODA} & \cellcolor[rgb]{0.980, 0.961, 0.937}
0.801/0.839  & \cellcolor[rgb]{0.980, 0.961, 0.937}
0.825/0.857 & 0.593/0.580 & 0.603/0.577 & 0.676/0.637& 0.541/0.679 & 0.646/0.530 & 0.649/0.536 & 0.542/0.588 & 0.490/0.580 &  {0.481/0.524} &  {0.478/0.549} \\ 
& \ct{RMA} / \ct{GMA} & \cellcolor[rgb]{0.980, 0.961, 0.937}
0.807/0.816  & \cellcolor[rgb]{0.980, 0.961, 0.937}
0.777/0.846 & 0.534/0.506 & 0.478/0.438 & 0.686/0.649 & 0.629/0.588 & 0.498/0.592 & 0.452/0.580 & 0.567/0.515 &  {0.498/0.438} &  {0.497/0.462} &  {0.486/0.456}  \\ 
& \ct{Untargeted} & \cellcolor[rgb]{0.980, 0.961, 0.937}
 {0.788}  & \cellcolor[rgb]{0.980, 0.961, 0.937}
0.790 & 0.546 & 0.577 & 0.698 & 0.655 & 0.507 & 0.486 & 0.614 & 0.646 & 0.518 & 0.528\\ 
& \ct{CIB} & \cellcolor[rgb]{0.980, 0.961, 0.937}
 {0.720}  & \cellcolor[rgb]{0.980, 0.961, 0.937}
0.756 & 0.631 & 0.676 & 0.583 & 0.632 & 0.522 & 0.548 & - & - & 0.235 & 0.280\\ 
& \ct{DC} & \cellcolor[rgb]{0.980, 0.961, 0.957}  {0.838} &  \cellcolor[rgb]{0.980, 0.961, 0.937}  {0.914} &
0.576 & 0.635 & 0.703 & 0.721 & 0.504 & 0.483 & 0.583 & 0.664 & 0.480 & 0.518 \\ 
& \cellcolor[rgb]{.95,.95,.95} Average & \cellcolor[rgb]{.95,.95,.95} \textbf{0.801} & \cellcolor[rgb]{.95,.95,.95} \textbf{0.824} & \cellcolor[rgb]{.95,.95,.95}0.567 & \cellcolor[rgb]{.95,.95,.95}0.569 & \cellcolor[rgb]{.95,.95,.95}0.662 & \cellcolor[rgb]{.95,.95,.95}0.635 & \cellcolor[rgb]{.95,.95,.95}0.543 & \cellcolor[rgb]{.95,.95,.95}0.533 & \cellcolor[rgb]{.95,.95,.95}0.568 & \cellcolor[rgb]{.95,.95,.95}0.553 & \cellcolor[rgb]{.95,.95,.95}0.457 & \cellcolor[rgb]{.95,.95,.95}0.471  \\ 
\midrule

\multirow{6}{*}{Faster-RCNN} & \ct{OGA} / \ct{ODA} & \cellcolor[rgb]{0.980, 0.961, 0.937}
0.757/0.854  & \cellcolor[rgb]{0.980, 0.961, 0.937}
0.770/0.856 & 0.601/0.610 & 0.601/0.624 & 0.716/0.670 & 0.591/0.606 & 0.631/0.497 & 0.636/0.546 & 0.542/0.588 & 0.490/0.580 & 0.498/0.481 & 0.482/0.469 \\ 
& \ct{RMA} / \ct{GMA} & \cellcolor[rgb]{0.980, 0.961, 0.937}
0.872/0.870  & \cellcolor[rgb]{0.980, 0.961, 0.937}
0.899/0.883 & 0.563/0.519 & 0.535/0.491 & 0.721/0.659 & 0.655/0.581 & 0.471/0.530 & 0.471/0.533 &  {0.567/0.512} & 0.498/0.439 & 0.495/0.491 & 0.491/0.503 \\ 
& \ct{Untargeted} & \cellcolor[rgb]{0.980, 0.961, 0.937}
 {0.869}  & \cellcolor[rgb]{0.980, 0.961, 0.937}
0.842 & 0.577 & 0.571 & 0.670 & 0.655 & 0.559 & 0.564 & 0.614 & 0.646 & 0.494 & 0.471 \\ 
& \ct{CIB} & \cellcolor[rgb]{0.980, 0.961, 0.937}
 {0.747}  & \cellcolor[rgb]{0.980, 0.961, 0.937}
0.750 & 0.638 & 0.682 & 0.613 & 0.602 & 0.585 & 0.603 & - & - & 0.262 & 0.329\\ 
& \ct{DC} & \cellcolor[rgb]{0.980, 0.961, 0.937}
 {0.935}  & \cellcolor[rgb]{0.980, 0.961, 0.937}
0.944 & 0.559 & 0.667 & 0.627 & 0.745 & 0.475 & 0.528 & 0.583 & 0.664 & 0.477 & 0.454 \\ 
& \cellcolor[rgb]{.95,.95,.95} Average & \cellcolor[rgb]{.95,.95,.95} \textbf{0.843} & \cellcolor[rgb]{.95,.95,.95} \textbf{0.849} & \cellcolor[rgb]{.95,.95,.95}0.581 & \cellcolor[rgb]{.95,.95,.95}0.597 & \cellcolor[rgb]{.95,.95,.95}0.668 & \cellcolor[rgb]{.95,.95,.95}0.633 & \cellcolor[rgb]{.95,.95,.95}0.535 & \cellcolor[rgb]{.95,.95,.95}0.554 & \cellcolor[rgb]{.95,.95,.95}0.568 & \cellcolor[rgb]{.95,.95,.95}0.553 & \cellcolor[rgb]{.95,.95,.95}0.457 & \cellcolor[rgb]{.95,.95,.95}0.457\\ 

\rowcolor{LightCyan}
\midrule
\multicolumn{14}{c}{\textbf{PASCAL VOC}~\citep{everingham2010pascal}} \\
\midrule
\multirow{6}{*}{YOLO} & \ct{OGA} / \ct{ODA} & \cellcolor[rgb]{0.980, 0.961, 0.937}
0.928/0.901  & \cellcolor[rgb]{0.980, 0.961, 0.937}
0.945/0.905 & 0.536/0.543 & 0.561/0.571 & 0.670/0.721 & 0.631/0.652 & 0.612/0.511 & 0.664/0.445 & 0.500/0.601 & 0.485/0.637 & 0.441/0.522 & 0.458/0.505\\ 
& \ct{RMA} / \ct{GMA} & \cellcolor[rgb]{0.980, 0.961, 0.937}
0.916/0.873 & \cellcolor[rgb]{0.980, 0.961, 0.937}
0.908/0.939 & 0.445/0.514 & 0.444/0.479 & 0.619/0.663 & 0.585/0.555 & 0.563/0.447 & 0.626/0.513 &  {0.562/0.477} & 0.584/0.502 & 0.515/0.508 & 0.495/0.489  \\ 
& \ct{Untargeted} & \cellcolor[rgb]{0.980, 0.961, 0.937}
 {0.800} & \cellcolor[rgb]{0.980, 0.961, 0.937}
0.858 & 0.487 & 0.497 & 0.706 & 0.701 & 0.551 & 0.503 & 0.644 & 0.687 & 0.482 & 0.490\\ 
& \ct{CIB} & \cellcolor[rgb]{0.980, 0.961, 0.937}
 {0.785}  & \cellcolor[rgb]{0.980, 0.961, 0.937}
0.767 & 0.610 & 0.656 & 0.618 & 0.624 & 0.567 & 0.598 & - & - & 0.273 & 0.332\\ 
& \ct{DC} & \cellcolor[rgb]{0.980, 0.961, 0.937}
 {0.936}  & \cellcolor[rgb]{0.980, 0.961, 0.937}
0.944 & 0.502 & 0.655 & 0.683 & 0.792 & 0.509 & 0.505 & 0.563 & 0.614 & 0.534 & 0.559\\ 
& \cellcolor[rgb]{.95,.95,.95} Average & \cellcolor[rgb]{.95,.95,.95} \textbf{0.877} & \cellcolor[rgb]{.95,.95,.95} \textbf{0.895} & \cellcolor[rgb]{.95,.95,.95}0.520 & \cellcolor[rgb]{.95,.95,.95}0.552 & \cellcolor[rgb]{.95,.95,.95}0.669 & \cellcolor[rgb]{.95,.95,.95}0.649 & \cellcolor[rgb]{.95,.95,.95}0.537 & \cellcolor[rgb]{.95,.95,.95}0.551 & \cellcolor[rgb]{.95,.95,.95}0.558 & \cellcolor[rgb]{.95,.95,.95}0.585 & \cellcolor[rgb]{.95,.95,.95}0.468 & \cellcolor[rgb]{.95,.95,.95}0.475 \\ 
\midrule

\multirow{6}{*}{DETR} & \ct{OGA} / \ct{ODA} & \cellcolor[rgb]{0.980, 0.961, 0.937}
0.775/0.823  & \cellcolor[rgb]{0.980, 0.961, 0.937}
0.799/0.831 & 0.524/0.515 & 0.547/0.520 & 0.703/0.608 & 0.627/0.637 & 0.610/0.519 & 0.660/0.467 & 0.500/0.601 & 0.485/0.637 & 0.483/0.505 & 0.497/0.521 \\ 
& \ct{RMA} / \ct{GMA} & \cellcolor[rgb]{0.980, 0.961, 0.937}
0.803/0.766  & \cellcolor[rgb]{0.980, 0.961, 0.937}
0.786/0.774 & 0.466/0.488 & 0.450/0.447 & 0.651/0.622 & 0.563/0.550 & 0.559/0.533 & 0.584/0.562 & {0.562/0.477} & 0.584/0.502  & 0.541/0.477 & 0.559/0.483 \\ 
& \ct{Untargeted} & \cellcolor[rgb]{0.980, 0.961, 0.937}
 {0.831}  & \cellcolor[rgb]{0.980, 0.961, 0.937}
0.820 & 0.567 & 0.572 & 0.655 & 0.705 & 0.647 & 0.692 & 0.644 & 0.687 & 0.489 & 0.476\\ 
& \ct{CIB} & \cellcolor[rgb]{0.980, 0.961, 0.937}
 {0.750}  & \cellcolor[rgb]{0.980, 0.961, 0.937}
0.767 & 0.632 & 0.683 & 0.591 & 0.587 & 0.575 & 0.569 & - & -& 0.275 & 0.346 \\ 
& \ct{DC} & \cellcolor[rgb]{0.980, 0.961, 0.937}
 {0.960} & \cellcolor[rgb]{0.980, 0.961, 0.937}
0.946 & 0.559 & 0.626 & 0.636 & 0.724 & 0.560 & 0.472 & 0.563 & 0.614 & 0.497 & 0.502 \\ 
& \cellcolor[rgb]{.95,.95,.95} Average & \cellcolor[rgb]{.95,.95,.95} \textbf{0.815} & \cellcolor[rgb]{.95,.95,.95} \textbf{0.818} & \cellcolor[rgb]{.95,.95,.95}0.536 & \cellcolor[rgb]{.95,.95,.95}0.549 & \cellcolor[rgb]{.95,.95,.95}0.638 & \cellcolor[rgb]{.95,.95,.95}0.628 & \cellcolor[rgb]{.95,.95,.95}0.572 & \cellcolor[rgb]{.95,.95,.95}0.572 & \cellcolor[rgb]{.95,.95,.95}0.558 & \cellcolor[rgb]{.95,.95,.95}0.585 & \cellcolor[rgb]{.95,.95,.95}0.467 & \cellcolor[rgb]{.95,.95,.95}0.483 \\

\rowcolor{LightCyan}
\midrule
\multicolumn{14}{c}{\textbf{Synthesized Traffic Signs}~\citep{shen2024django}} \\
\midrule

\multirow{6}{*}{YOLO} & \ct{OGA} / \ct{ODA} & \cellcolor[rgb]{0.980, 0.961, 0.937}
0.917/0.885  & \cellcolor[rgb]{0.980, 0.961, 0.937}
0.942/0.872 & 0.581/0.591 & 0.581/0.560 & 0.693/0.640 & 0.665/0.664 & 0.671/0.442 & 0.645/0.467 &  0.504/0.606 &   0.529/0.561 &  0.473/0.453&  0.480/0.446 \\ 
& \ct{RMA} / \ct{GMA} & \cellcolor[rgb]{0.980, 0.961, 0.937}
0.957/0.928  & \cellcolor[rgb]{0.980, 0.961, 0.937}
0.944/0.936 & 0.515/0.485 & 0.459/0.490 & 0.678/0.690 & 0.605/0.549 & 0.494/0.511 & 0.523/0.579 & {0.562/0.522} & 0.506/0.424 &  0.505/0.481 &  0.517/0.472\\ 
& \ct{Untargeted} & \cellcolor[rgb]{0.980, 0.961, 0.937}
 {0.857} &  \cellcolor[rgb]{0.980, 0.961, 0.937}
0.903 & 0.550 & 0.501 & 0.703 & 0.645 & 0.467 & 0.441 & 0.639 & 0.613 & 0.517 & 0.494\\ 
& \ct{CIB} & \cellcolor[rgb]{0.980, 0.961, 0.937}
 {0.825}  & \cellcolor[rgb]{0.980, 0.961, 0.937}
0.853 & 0.628 & 0.649 & 0.655 & 0.673 & 0.548 & 0.591 &  - &  - & 0.236 & 0.354\\ 
& \ct{DC} & \cellcolor[rgb]{0.980, 0.961, 0.937}
 {0.926} & \cellcolor[rgb]{0.980, 0.961, 0.937}
0.941 & 0.557 & 0.689 & 0.701 & 0.754 & 0.518 & 0.523 & 0.590 & 0.543 & 0.511 & 0.532\\ 
& \cellcolor[rgb]{.95,.95,.95} Average & \cellcolor[rgb]{.95,.95,.95} \textbf{0.899} & \cellcolor[rgb]{.95,.95,.95} \textbf{0.914} & \cellcolor[rgb]{.95,.95,.95}0.558 & \cellcolor[rgb]{.95,.95,.95}0.561 & \cellcolor[rgb]{.95,.95,.95}0.680 & \cellcolor[rgb]{.95,.95,.95}0.651 & \cellcolor[rgb]{.95,.95,.95}0.522 & \cellcolor[rgb]{.95,.95,.95}0.538 & \cellcolor[rgb]{.95,.95,.95}0.571 & \cellcolor[rgb]{.95,.95,.95}0.529  & \cellcolor[rgb]{.95,.95,.95}0.454  & \cellcolor[rgb]{.95,.95,.95}0.471 
 \\ 
\midrule
\bottomrule[1pt]
\end{tabular}
\end{threeparttable}

}
\label{Tab:Exp1}
\end{table*}

\noindent\textbf{Defense Settings.}
We compare \ourmethod with \ct{Detector Cleanse}~\citep{chan2022baddet}, the only current black-box TTSD tailored for OD. We also include four black-box defenses applicable to OD: \ct{Strip}~\citep{gao2019strip}, \ct{FreqDetector}~\citep{zeng2021rethinking}, \ct{TeCo}~\citep{liu2023detecting}, and \ct{SCALE-UP}~\citep{guo2022scale}. We follow the configurations recommended in their papers for a fair comparison. 

\noindent\textbf{Evaluation Metrics.}
We evaluate using two metrics: F1 score (balancing precision and recall) and AUROC (measuring overall performance across different thresholds). 

\subsection{Defense Performance}
\label{sec:exp_performance}
\noindent\textbf{Effectiveness Study.} 
In Tab.~\ref{Tab:Exp1}, \ourmethod consistently achieves promising performance across seven backdoor attacks, three datasets, three model architectures, and a total of 42 backdoored models.
To further assess \ourmethod's stability, we calculate performance standard deviations across different backdoors on the same model and dataset, finding low  standard variances (F1: $0.880 \pm 0.040$, AUROC: $0.897 \pm 0.041$), which demonstrates its robust consistency.

\noindent\textbf{Comparison Study.} 
Tab.~\ref{Tab:Exp1} also shows that \ourmethod comprehensively outperformed its five black-box competitors: \ct{TeCo}, \ct{Strip}, \ct{FreqDetector}, \ct{SCALE-UP}, and the SOTA \ct{Detector Cleanse} designed for OD task. The first four fail completely due to shifts in task context. Even though \ct{Detector Cleanse} is not attack-agnostic, \ourmethod still significantly surpasses it in F1 score by $\sim 30\%$. Further analysis is deferred to Appendix D.2.

\begin{figure}[!t]  
\setlength{\belowcaptionskip}{-0.6cm} 
\centering          
\includegraphics[width=0.5\textwidth]{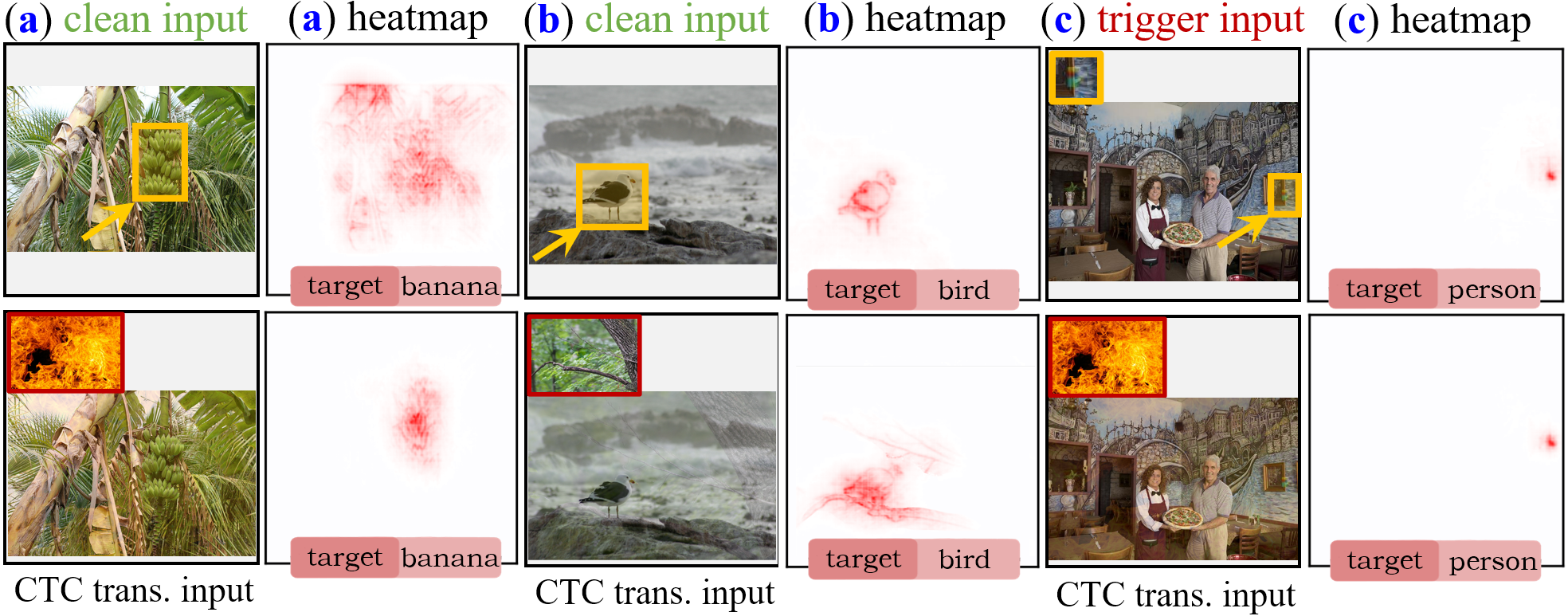}    
\caption{High resolution, object-specific explanations from L-CRP~\cite{dreyer2023revealing}.}       
\label{Fig:visual}  
\end{figure}

\noindent\textbf{A Closer Look to the Effectiveness of \ourmethod.} 
We use SOTA \textit{eXplainable Artificial Intelligence} (XAI) method L-CRP~\cite{dreyer2023revealing} to attain object-class-specific explanations in this multi-object context. We offer three illustrative examples: Fig.~\ref{Fig:visual}\textcolor{blue}{(\textbf{a})} shows that the detection of a banana relies heavily on surrounding green plants. \ourmethod’s background transformation (\ie introducing a contrasting fire scene) breaks this context bias, shifting reliance to the banana’s intrinsic features.
In Fig.~\ref{Fig:visual}\textcolor{blue}{(\textbf{b})}, the bird’s detection initially depends on its own features, but after transformation, new contextual dependencies emerge. In both cases, confidence scores shift (decrease for the banana, increasing for the bird). Lastly, Fig.~\ref{Fig:visual}\textcolor{blue}{(\textbf{c})} shows an \textit{FP-inducing} trigger, showing always input-independent detection.

\subsection{Ablation Study}

\begin{figure}[t]
\setlength{\belowcaptionskip}{-0.3cm}
\centering
    \begin{subfigure}{0.23\textwidth}
        \includegraphics[width=\textwidth]{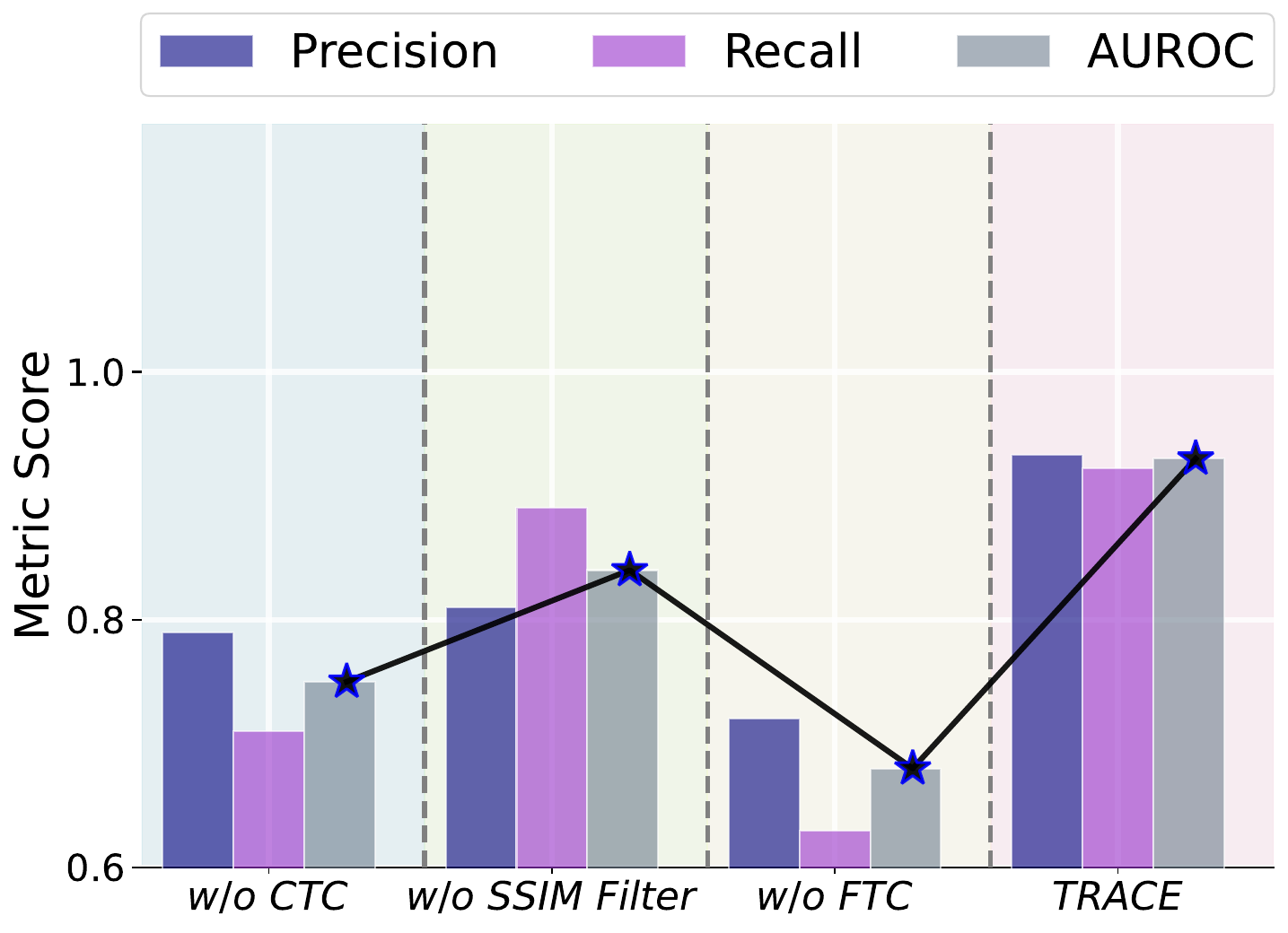} 
        \caption{Defending \ct{RMA} (\textit{hybrid attack})}
        \label{Fig.max_cifar10}
    \end{subfigure}
    \begin{subfigure}{0.23\textwidth}
        \includegraphics[width=\textwidth]{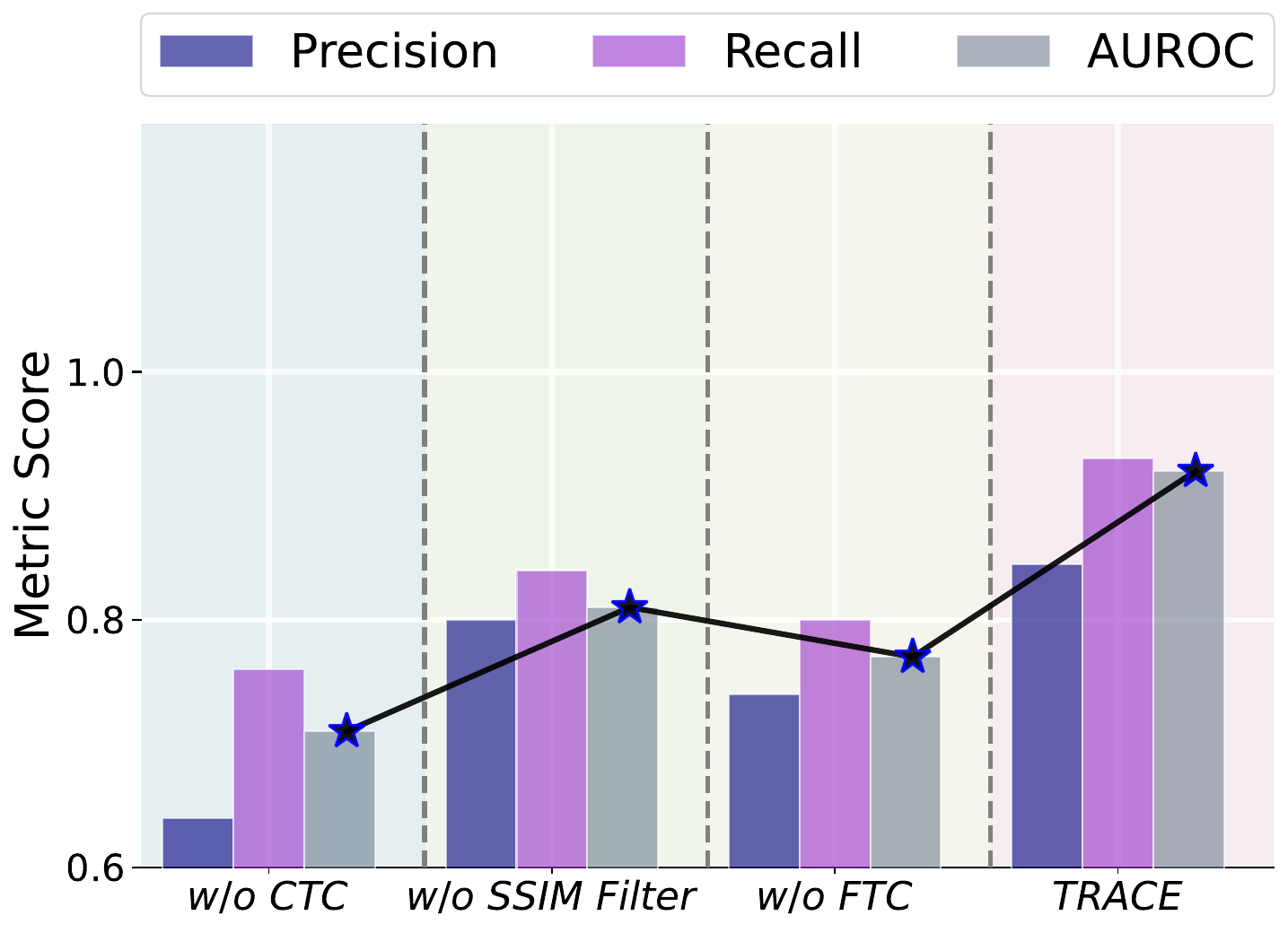} 
        \caption{Defending \ct{GMA} (\textit{hybrid attack})}
        \label{Fig.max_tiny}
    \end{subfigure}
\caption{\textbf{(Ablation Study)} Effect of different modules.}
\label{fig:ablation}
\end{figure}

\noindent \textbf{Effect of Modules.} 
\ourmethod incorporates three fundamental designs: contextual transformation, focal transformation, and SSIM-filter post-processing. We evaluate their individual contributions by isolating each component's impact.
Fig.~\ref{fig:ablation}\textcolor{blue}{(\textbf{a-b})} presents results for defending against \ct{RMA} and \ct{GMA} (two hybrid attacks), with each cluster of three bars representing an ablation case. As expected, omitting any component impairs performance, underscoring the complementary and essential roles of each module.


\noindent \textbf{Effect of Thresholds $\gamma$.} 
\ourmethod maps an input sample $\bm{x}$ to a linearly separable space, defenders classify it based on a threshold $\gamma$. 
In our evaluation, we set an empirical threshold for \ourmethod. Even in the worst case where no information is available to estimate an appropriate threshold, by setting $\gamma=0$, \ourmethod can still get an competitive average F1 score, still surpassing \ct{Detector Cleanse}'s performance at optimal thresholds.

\begin{figure}[t]
\setlength{\belowcaptionskip}{-0.3cm}
\centering
    \begin{subfigure}{0.156\textwidth}
        \includegraphics[width=\textwidth]{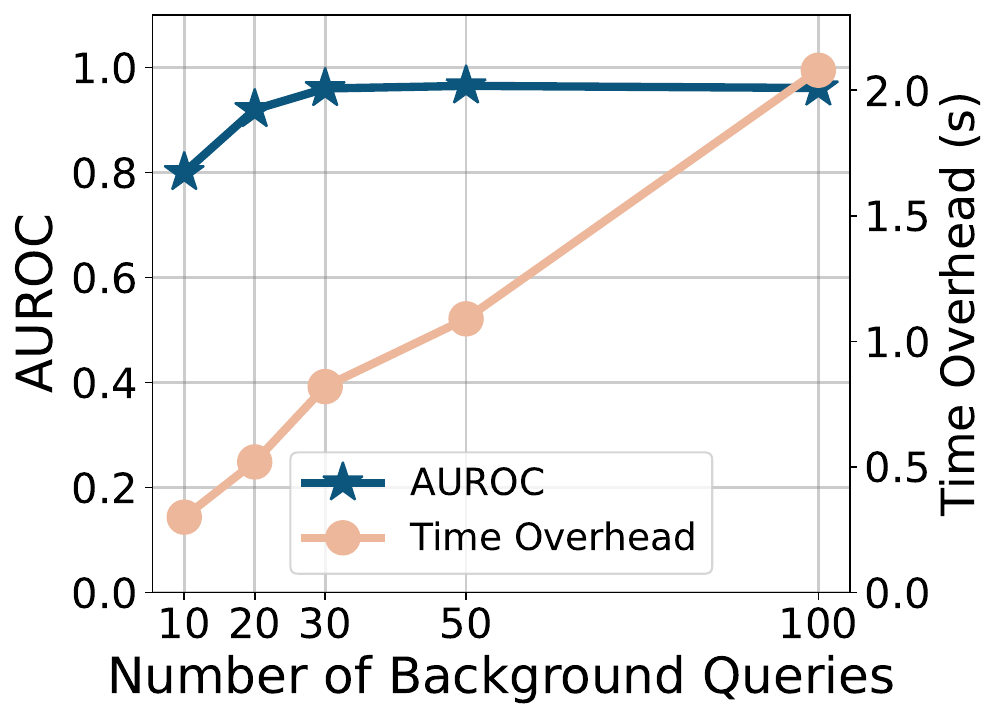} 
        \caption{Background queries}
        \label{Fig:hyb-a}
    \end{subfigure}
    \begin{subfigure}{0.156\textwidth}
        \includegraphics[width=\textwidth]{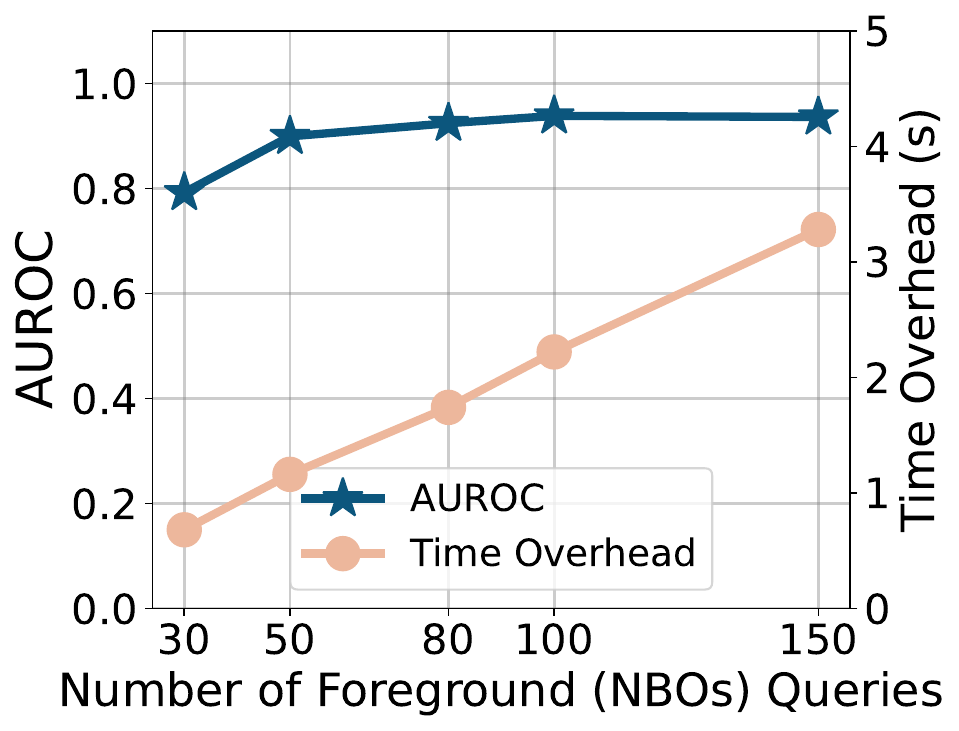} 
       \caption{Foreground queries}
        \label{Fig:hyb-b}
    \end{subfigure}
    \begin{subfigure}{0.156\textwidth}
        \includegraphics[width=\textwidth]{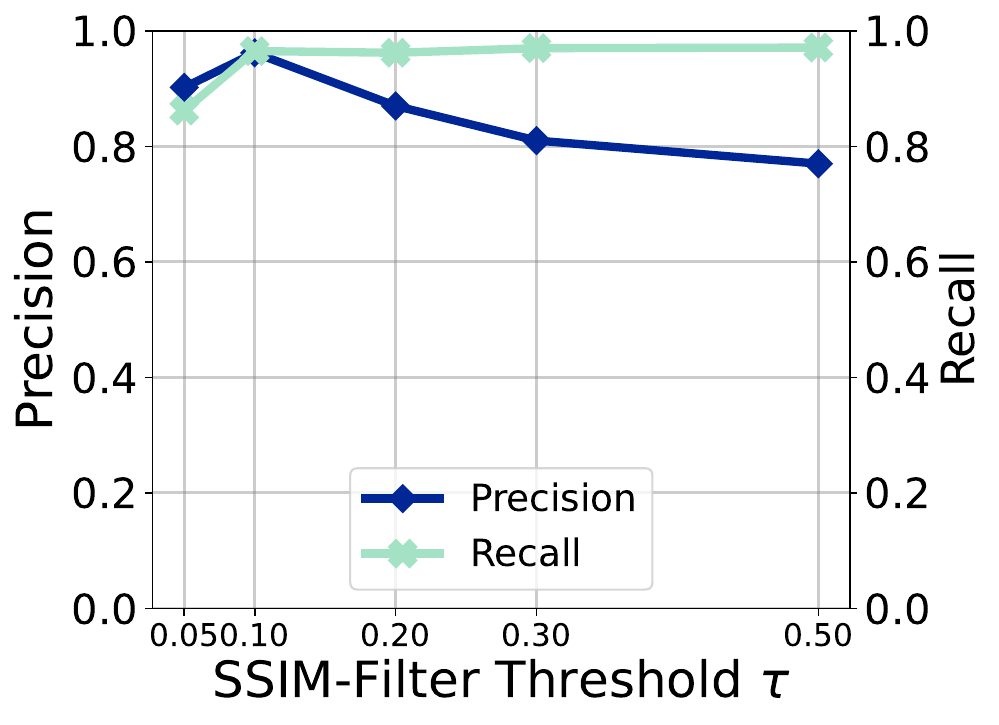} 
        \caption{$\tau$ value}
        \label{Fig:hyb-c}
    \end{subfigure}
\caption{\textbf{(Ablation Study)} Effect of background, foreground, and $\tau$.}
\label{Fig:hyb}
\end{figure}

\noindent \textbf{Effect of Hyperparameters $b, f, \tau$.} 
As shown in Fig.~\ref{Fig:hyb}(\textcolor{blue}{\textbf{a}}-\textcolor{blue}{\textbf{b}}), performance improves with increasing $b$ and $f$, albeit with higher time overhead. A balance is reached at $b=30$ and $f=50$, achieving strong performance with reasonable time efficiency. 
Fig.~\ref{Fig:hyb}(\textcolor{blue}{\textbf{c}}) shows the impact of different $\tau$ when distinguishing between NBOs and triggers. We select the $\tau=0.1$ as the optimal balance point.




\begin{figure}[t]
\setlength{\belowcaptionskip}{-0.3cm}
\centering
    \begin{subfigure}{0.21\textwidth}
        \includegraphics[width=\textwidth]{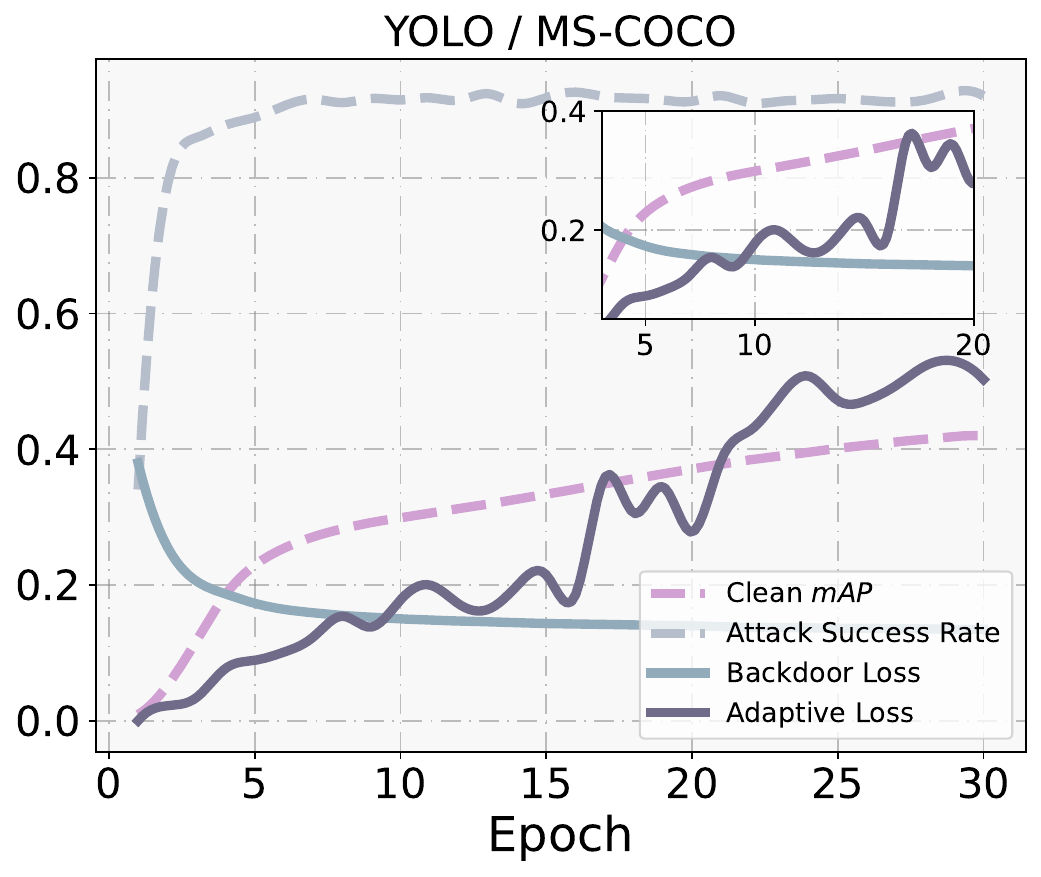} 
        \caption{\ct{OGA} backdoor attack}
        \label{Fig.max_cifar10}
    \end{subfigure}
    \begin{subfigure}{0.21\textwidth}
        \includegraphics[width=\textwidth]{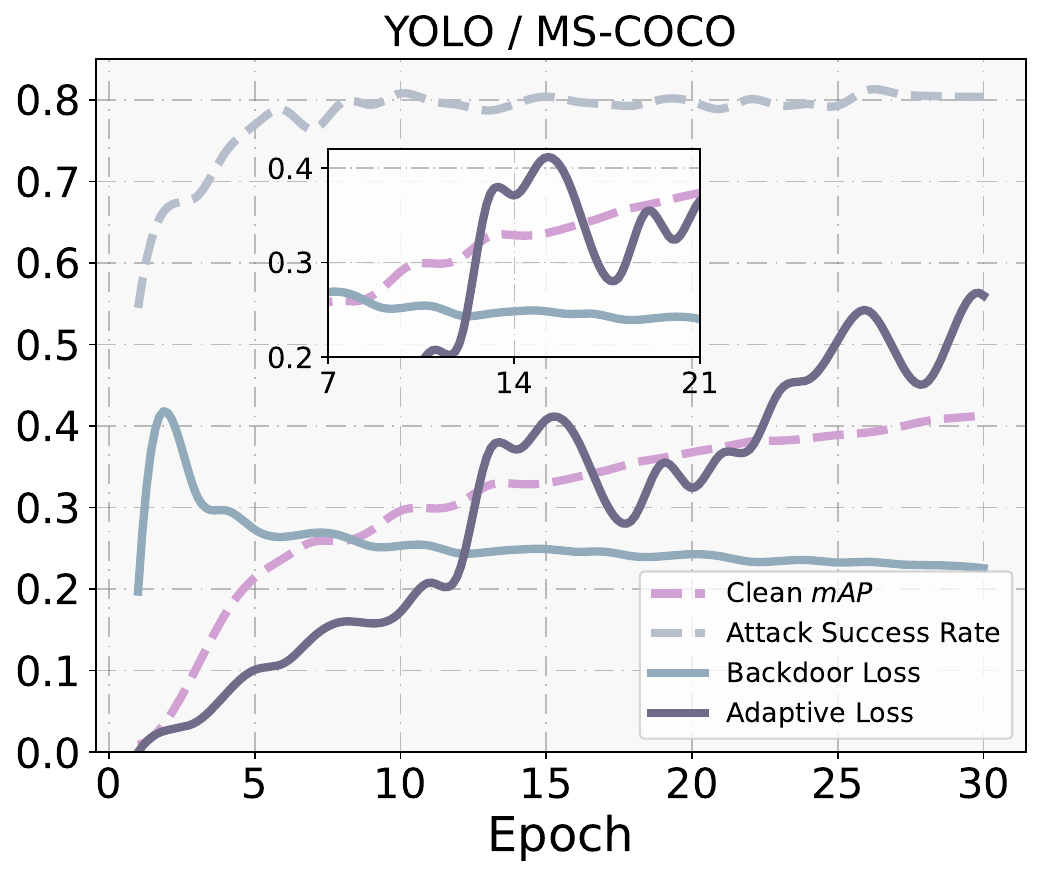} 
        \caption{\ct{ODA} backdoor attack}
        \label{Fig.max_tiny}
    \end{subfigure}
\caption{\textbf{(Adaptive attack)} Visualization of clean mAP (mean Average Precision), backdoor loss, attack success rate, and adaptive loss in backdoor training. Note the \textit{negative correlation}: as backdoor loss decreases, adaptive loss rises correspondingly.}
\label{fig:loss}
\end{figure}

\subsection{Resistance to Potential Adaptive Attacks} 

\begin{table}[!t]
\centering
\caption{\textbf{(Adaptive attack)} \ourmethod against the omniscient adaptive attack.}
\label{tab:ada}
\setlength{\tabcolsep}{4pt}
\resizebox{0.48\textwidth}{!}{%
\begin{tabular}{lcccccccccccc} 
\toprule[1.2pt]
 $\lambda$$\rightarrow$ & \multicolumn{3}{c}{$5\times10^{-4}$} & \multicolumn{3}{c}{$1\times10^{-4}$} & \multicolumn{3}{c}{$5\times10^{-5}$} & \multicolumn{3}{c}{$1\times10^{-5}$} \\
  \cmidrule{2-13}
 Attacks$\downarrow$ & F1 & AUROC & ASR & F1 & AUROC & ASR & F1 & AUROC & ASR & F1 & AUROC  & ASR\\ 
\midrule
\ct{OGA} & 0.581& 0.620& 0.733& 0.766& 0.747& 0.843& 0.916& 0.879& 0.944& 0.937 & 0.942& 0.975\\
\ct{ODA} & 0.606& 0.594& 0.583& 0.608& 0.576& 0.752& 0.820& 0.794&  0.841& 0.840& 0.851& 0.893\\ 
\ct{RMA} & 0.617&0.614 & 0.681& 0.748& 0.702& 0.810& 0.867& 0.886& 0.933& 0.938& 0.929& 0.947\\
\bottomrule[1.2pt]
\end{tabular}
}
\end{table}

We further challenge stronger adversaries in the worst-case scenario, assuming full knowledge of \ourmethod. 
Recall the dual targets of backdoored models in Sec.~\ref{Sec:formulation}, where the loss for training such models is $\mathcal{J}_{bd}=\sum_{i=1}^{\left|\mathcal{D}_b\right|} \mathcal{L}\left(\mathcal{F}\left(\boldsymbol{x}_i\right), y_i\right)+\sum_{j=1}^{\left|\mathcal{D}_p\right|} \mathcal{L}\left(\mathcal{F}\left(\boldsymbol{x}_j\right), y_t\right),$
where $\mathcal{L}(\cdot)$ is the cross entropy loss. We develop an adaptive loss to bypass it:
\small
\begin{equation}
    \begin{split}
        &\mathcal{J}_{adap} =  \sum_{i=1}^N \sum_{\delta_j}^{\mathcal{U}(B)}  MSE\left( \Delta_{\mathcal{B}}^{\mathrm{Var}}(\mathbb{F}_{\theta}, \bm{x}_i),\Delta_{\mathcal{B}}^{\mathrm{Var}}(\mathbb{F}_{\theta}, \bm{x}_i\oplus t)
        \right) + \\
        &  \sum_{i=1}^N MSE\left( \Delta_{\mathcal{F}}^{\mathrm{Var}}(\mathbb{F}_\theta, \bm{x}_i, o_{\mathrm{nbo}}), \Delta_{\mathcal{F}}^{\mathrm{Var}}(\mathbb{F}_\theta, \bm{x}_i\oplus t , o_{\mathrm{nbo}}) \right) \nonumber
    \end{split}
\end{equation}
\normalsize 
where the two terms respectively force the model to maintain the same contextual and focal transformation consistency for clean and poisoned samples, aligning with the \ourmethod's inference logic. We monitor the adaptive loss \textit{without derivative} during training. As backdoor loss decreases, adaptive loss rises (\ie a negative correlation, see Fig.~\ref{fig:loss}), suggesting that success in the dual-target loss may alter the model’s transformation consistency. Appendix E.3 offers a reasonable hypothesis. 
Subsequently, we integrate $\mathcal{J}_{adap}$ with the vanilla loss $\mathcal{J}_{bd}$ to formulate the overall loss as $\mathcal{J}=\mathcal{J}_{bd}+\lambda\mathcal{J}_{adap}$, where $\lambda$ is a weighting factor. We then optimize $\mathbb{F}_\theta$ by minimizing $\mathcal{J}$. Tab.~\ref{tab:ada} shows that while adaptive attacks can partially evade \ourmethod, they significantly sacrifice attack performance under $\mathcal{J}_{adap}$. This is due to the conflict between $\mathcal{J}_{adap}$ and $\mathcal{J}_{bd}$, with its failure once again confirming our sustained robustness.

\section{Conclusions, Limitations, and Future Work}
We present \ourmethod, a brand-new TTSD for object detection. \textit{\ourmethod is practical}, requiring only black-box model access and no training data. \textit{\ourmethod is universal}, demonstrating attack-agnostic stability against diverse attacks. \textit{\ourmethod is effective}, with extensive experiments on 63 backdoored models validating its performance. However, \ourmethod's limitations include increased time overhead due to varied transformations and reliance on auxiliary public data. Thus, efficient zero-shot detection is left for our future work.

\section*{Acknowledgement}
This work is supported by the National Key Research and Development Program of China (Grant No. 2023YFB4503400), and the National Natural Science Foundation of China (Grant Nos. 62450064, 62322205). Shengshan Hu is the corresponding author.

{
    \small
    \bibliographystyle{ieeenat_fullname}
    \bibliography{main}

@inproceedings{ma2022beatrix,   
title={The "Beatrix'' Resurrections: Robust Backdoor Detection via Gram Matrices},  
author={Ma, Wanlun and Wang, Derui and Sun, Ruoxi and Xue, Minhui and Wen, Sheng and Xiang, Yang},  
booktitle = {Proceedings of the Network and Distributed System Security Symposium (NDSS'23)},  
year={2023} 
}

@inproceedings{tang2021demon,
  title={Demon in the variant: Statistical analysis of DNNs for robust backdoor contamination detection},
  author={Tang, Di and Wang, XiaoFeng and Tang, Haixu and Zhang, Kehuan},
  booktitle={Proceedings of the 30th USENIX Security Symposium (USENIX Security'21)},
  pages={1541--1558},
  year={2021}
}

@inproceedings{carion2020end,   title={End-to-end object detection with transformers},   author={Carion, Nicolas and Massa, Francisco and Synnaeve, Gabriel and Usunier, Nicolas and Kirillov, Alexander and Zagoruyko, Sergey},   booktitle={Proceedings of the European Conference on Computer Vision (ECCV'20)},   pages={213--229},   year={2020} }

@inproceedings{chan2022baddet,   title={Baddet: Backdoor attacks on object detection},   author={Chan, Shih-Han and Dong, Yinpeng and Zhu, Jun and Zhang, Xiaolu and Zhou, Jun},   booktitle={Proceedings of the European Conference on Computer Vision (ECCV'22)},   pages={396--412},   year={2022},   organization={Springer} }

@inproceedings{luo2023untargeted,   title={Untargeted backdoor attack against object detection},   author={Luo, Chengxiao and Li, Yiming and Jiang, Yong and Xia, Shu-Tao},   booktitle={Proceedings of the IEEE International Conference on Acoustics, Speech and Signal Processing (ICASSP'23)},   
pages={1--5}, 
year={2023},  
organization={IEEE} 
}

@article{ma2022dangerous,
  title={Dangerous cloaking: Natural trigger based backdoor attacks on object detectors in the physical world},
  author={Ma, Hua and Li, Yinshan and Gao, Yansong and Abuadbba, Alsharif and Zhang, Zhi and Fu, Anmin and Kim, Hyoungshick and Al-Sarawi, Said F and Surya, Nepal and Abbott, Derek},
  journal={arXiv preprint arXiv:2201.08619},
  year={2022}
}

@inproceedings{wu2022just,   title={Just rotate it: Deploying backdoor attacks via rotation transformation},   author={Wu, Tong and Wang, Tianhao and Sehwag, Vikash and Mahloujifar, Saeed and Mittal, Prateek},   booktitle={Proceedings of the 15th ACM Workshop on Artificial Intelligence and Security},   pages={91--102},   year={2022} }

@article{cheng2023backdoor,   title={Backdoor Attack against Object Detection with Clean Annotation},   author={Cheng, Yize and Hu, Wenbin and Cheng, Minhao},   journal={arXiv preprint arXiv:2307.10487},   year={2023} }

@inproceedings{chen2022clean,   title={Clean-image backdoor: Attacking multi-label models with poisoned labels only},   author={Chen, Kangjie and Lou, Xiaoxuan and Xu, Guowen and Li, Jiwei and Zhang, Tianwei},   booktitle={Proceedings of the Eleventh International Conference on Learning Representations (ICLR'22)},   year={2022} }

@inproceedings{lin2020composite,   title={Composite backdoor attack for deep neural network by mixing existing benign features},   author={Lin, Junyu and Xu, Lei and Liu, Yingqi and Zhang, Xiangyu},   booktitle={Proceedings of the 2020 ACM SIGSAC Conference on Computer and Communications Security},   pages={113--131},   year={2020} }

@inproceedings{liu2023detecting,   title={Detecting backdoors during the inference stage based on corruption robustness consistency},   author={Liu, Xiaogeng and Li, Minghui and Wang, Haoyu and Hu, Shengshan and Ye, Dengpan and Jin, Hai and Wu, Libing and Xiao, Chaowei},   booktitle={Proceedings of the IEEE/CVF Conference on Computer Vision and Pattern Recognition (CVPR'23)},   pages={16363--16372},   year={2023} }

@inproceedings{zeng2021rethinking,   
title={Rethinking the backdoor attacks' triggers: A frequency perspective},  
author={Zeng, Yi and Park, Won and Mao, Z Morley and Jia, Ruoxi}, 
booktitle={Proceedings of the IEEE/CVF international conference on computer vision (ICCV'21)},   pages={16473--16481},   
year={2021} 
}

@inproceedings{gao2019strip,   title={Strip: A defence against trojan attacks on deep neural networks},   author={Gao, Yansong and Xu, Change and Wang, Derui and Chen, Shiping and Ranasinghe, Damith C and Nepal, Surya},   booktitle={Proceedings of the 35th annual computer security applications conference (ACSAC'19)},   pages={113--125},   year={2019} }

@article{shen2024django,   title={Django: Detecting trojans in object detection models via gaussian focus calibration},   author={Shen, Guangyu and Cheng, Siyuan and Tao, Guanhong and Zhang, Kaiyuan and Liu, Yingqi and An, Shengwei and Ma, Shiqing and Zhang, Xiangyu},   journal={Proceedings of the Advances in Neural Information Processing Systems (NeurIPS'24)},   volume={36},   year={2024} }

@inproceedings{cheng2024odscan,   title={ODSCAN: Backdoor Scanning for Object Detection Models},   author={Cheng, Siyuan and Shen, Guangyu and Tao, Guanhong and Zhang, Kaiyuan and Zhang, Zhuo and An, Shengwei and Xu, Xiangzhe and Liu, Yingqi and Ma, Shiqing and Zhang, Xiangyu},   booktitle={Proceedings of the IEEE Symposium on Security and Privacy (SP'24)},   pages={119--119},   year={2024}}

@article{udeshi2022model,   title={Model agnostic defence against backdoor attacks in machine learning},   author={Udeshi, Sakshi and Peng, Shanshan and Woo, Gerald and Loh, Lionell and Rawshan, Louth and Chattopadhyay, Sudipta},   journal={IEEE Transactions on Reliability},  
pages={880--895},  
year={2022}
}

@article{ren2015faster,   title={Faster r-cnn: Towards real-time object detection with region proposal networks},   author={Ren, Shaoqing and He, Kaiming and Girshick, Ross and Sun, Jian},   journal={Proceedings of the Advances in Neural Information Processing Systems (NeurIPS'15)},   volume={28},   year={2015} }

@inproceedings{redmon2016you,   
title={You only look once: Unified, real-time object detection},   
author={Redmon, Joseph and Divvala, Santosh and Girshick, Ross and Farhadi, Ali},   booktitle={Proceedings of the IEEE conference on computer vision and pattern recognition (CVPR'16)},  
pages={779--788},  
year={2016} 
}

@article{everingham2010pascal,   title={The pascal visual object classes (voc) challenge},   author={Everingham, Mark and Van Gool, Luc and Williams, Christopher KI and Winn, John and Zisserman, Andrew},   journal={International journal of computer vision},   pages={303--338},   year={2010}}

@inproceedings{guo2022scale,  
title={SCALE-UP: An Efficient Black-box Input-level Backdoor Detection via Analyzing Scaled Prediction Consistency},   author={Guo, Junfeng and Li, Yiming and Chen, Xun and Guo, Hanqing and Sun, Lichao and Liu, Cong},   booktitle={Proceedings of the Eleventh International Conference on Learning Representations (ICLR'23)},  
year={2023} 
}

@inproceedings{wang2016cnn,   title={Cnn-rnn: A unified framework for multi-label image classification},   author={Wang, Jiang and Yang, Yi and Mao, Junhua and Huang, Zhiheng and Huang, Chang and Xu, Wei},   booktitle={Proceedings of the IEEE conference on computer vision and pattern recognition (CVPR'16)},   pages={2285--2294},   year={2016} }

@inproceedings{yazici2020orderless,   title={Orderless recurrent models for multi-label classification},   author={Yazici, Vacit Oguz and Gonzalez-Garcia, Abel and Ramisa, Arnau and Twardowski, Bartlomiej and Weijer, Joost van de},   booktitle={Proceedings of the IEEE/CVF Conference on Computer Vision and Pattern Recognition (CVPR'20)},   pages={13440--13449},   year={2020} }

@inproceedings{chen2019multi,   title={Multi-label image recognition with graph convolutional networks},   author={Chen, Zhao-Min and Wei, Xiu-Shen and Wang, Peng and Guo, Yanwen},   booktitle={Proceedings of the IEEE/CVF conference on computer vision and pattern recognition (CVPR'19)},   pages={5177--5186},   year={2019} }

@inproceedings{durand2019learning,   title={Learning a deep convnet for multi-label classification with partial labels},   author={Durand, Thibaut and Mehrasa, Nazanin and Mori, Greg},   booktitle={Proceedings of the IEEE/CVF conference on computer vision and pattern recognition (CVPR'19)},   pages={647--657},   year={2019} }

@inproceedings{zhao2021transformer,   title={Transformer-based dual relation graph for multi-label image recognition},   author={Zhao, Jiawei and Yan, Ke and Zhao, Yifan and Guo, Xiaowei and Huang, Feiyue and Li, Jia},   booktitle={Proceedings of the IEEE/CVF international conference on computer vision (ICCV'21)},   pages={163--172},   year={2021} }

@article{shortcut,
  title={Shortcut learning in deep neural networks},
  author={Geirhos, Robert and Jacobsen, J{\"o}rn-Henrik and Michaelis, Claudio and Zemel, Richard and Brendel, Wieland and Bethge, Matthias and Wichmann, Felix A},
  journal={Nature Machine Intelligence},
  pages={665--673},
  year={2020}
}

@inproceedings{geirhos2018imagenet,
  title={ImageNet-trained CNNs are biased towards texture; increasing shape bias improves accuracy and robustness},
  author={Geirhos, Robert and Rubisch, Patricia and Michaelis, Claudio and Bethge, Matthias and Wichmann, Felix A and Brendel, Wieland},
  booktitle={Proceedings of the International Conference on Learning Representations (ICLR'18)},
  year={2018}
}

@inproceedings{muhammad2020eigen,
  title={Eigen-cam: Class activation map using principal components},
  author={Muhammad, Mohammed Bany and Yeasin, Mohammed},
  booktitle={Proceedings of the international joint conference on neural networks (IJCNN)},
  pages={1--7},
  year={2020},
}

@inproceedings{liu2022contextual,
  title={Contextual debiasing for visual recognition with causal mechanisms},
  author={Liu, Ruyang and Liu, Hao and Li, Ge and Hou, Haodi and Yu, TingHao and Yang, Tao},
  booktitle={Proceedings of the IEEE/CVF Conference on Computer Vision and Pattern Recognition (CVPR'22)},
  pages={12755--12765},
  year={2022}
}

@article{hesamifard2018privacy,
  title={Privacy-preserving Machine Learning as a Service},
  author={Hesamifard, Ehsan and Takabi, Hassan and Ghasemi, Mehdi and Wright, Rebecca N},
  journal={Proceedings on Privacy Enhancing Technologies},
  volume={2018},
  number={3},
  pages={123--142},
  year={2018},
  publisher={Privacy Enhancing Technologies Symposium Advisory Board}
}

@inproceedings{ribeiro2015mlaas,
  title={Mlaas: Machine learning as a service},
  author={Ribeiro, Mauro and Grolinger, Katarina and Capretz, Miriam AM},
  booktitle={Proceedings of the IEEE 14th international conference on machine learning and applications (ICMLA'15)},
  pages={896--902},
  year={2015},
  organization={IEEE}
}

@article{xiao2020noise, 
title={Noise or signal: The role of image backgrounds in object recognition},   author={Xiao, Kai and Engstrom, Logan and Ilyas, Andrew and Madry, Aleksander},   
journal={arXiv preprint arXiv:2006.09994},
year={2020} 
}

@article{geirhos2020shortcut,  
title={Shortcut learning in deep neural networks},  
author={Geirhos, Robert and Jacobsen, J{\"o}rn-Henrik and Michaelis, Claudio and Zemel, Richard and Brendel, Wieland and Bethge, Matthias and Wichmann, Felix A},  
journal={Nature Machine Intelligence},   volume={2}, 
number={11}, 
pages={665--673},  
year={2020}, 
}

@article{anders2022finding,
  title={Finding and removing clever hans: Using explanation methods to debug and improve deep models},
  author={Anders, Christopher J and Weber, Leander and Neumann, David and Samek, Wojciech and M{\"u}ller, Klaus-Robert and Lapuschkin, Sebastian},
  journal={Information Fusion},
  volume={77},
  pages={261--295},
  year={2022},
}

@inproceedings{hou2024ibd,
  title={IBD-PSC: Input-level Backdoor Detection via Parameter-oriented Scaling Consistency},
  author={Hou, Linshan and Feng, Ruili and Hua, Zhongyun and Luo, Wei and Zhang, Leo Yu and Li, Yiming},
  booktitle={Proceedings of the Forty-first International Conference on Machine Learning (ICML'24)},
  year={2024}
}

@inproceedings{dreyer2023revealing,  
title={Revealing hidden context bias in segmentation and object detection through concept-specific explanations}, 
author={Dreyer, Maximilian and Achtibat, Reduan and Wiegand, Thomas and Samek, Wojciech and Lapuschkin, Sebastian},  
booktitle={Proceedings of the IEEE/CVF Conference on Computer Vision and Pattern Recognition (CVPR'23)},  
pages={3828--3838},  
year={2023} 
}

@inproceedings{doan2020februus,   
title={Februus: Input purification defense against trojan attacks on deep neural network systems},   author={Doan, Bao Gia and Abbasnejad, Ehsan and Ranasinghe, Damith C},  
booktitle={Proceedings of the 36th Annual Computer Security Applications Conference},  
pages={897--912}, 
year={2020} 
}

@inproceedings{guan2024backdoor,
  title={Backdoor Defense via Test-Time Detecting and Repairing},
  author={Guan, Jiyang and Liang, Jian and He, Ran},
  booktitle={Proceedings of the IEEE/CVF Conference on Computer Vision and Pattern Recognition (CVPR'24)},
  pages={24564--24573},
  year={2024}
}

@inproceedings{wang2023does,   
title={Does physical adversarial example really matter to autonomous driving? towards system-level effect of adversarial object evasion attack}, 
author={Wang, Ningfei and Luo, Yunpeng and Sato, Takami and Xu, Kaidi and Chen, Qi Alfred}, 
booktitle={Proceedings of the IEEE/CVF International Conference on Computer Vision (ICCV'23)},  
pages={4412--4423},
year={2023} 
}

@article{wang2024deep,
  title={Deep intra-image contrastive learning for weakly supervised one-step person search},
  author={Wang, Jiabei and Pang, Yanwei and Cao, Jiale and Sun, Hanqing and Shao, Zhuang and Li, Xuelong},
  journal={Pattern Recognition},
  volume={147},
  pages={110047},
  year={2024},
}

@inproceedings{lin2014microsoft,  
title={Microsoft coco: Common objects in context},   
author={Lin, Tsung-Yi and Maire, Michael and Belongie, Serge and Hays, James and Perona, Pietro and Ramanan, Deva and Doll{\'a}r, Piotr and Zitnick, C Lawrence},   booktitle={Computer Vision--ECCV 2014: 13th European Conference, Zurich, Switzerland, September 6-12, 2014, Proceedings, Part V 13},  
pages={740--755}, 
year={2014},   
organization={Springer}
}

@inproceedings{zhang2024detector,   
title={Detector collapse: Backdooring object detection to catastrophic overload or blindness},   author={Zhang, Hangtao and Hu, Shengshan and Wang, Yichen and Zhang, Leo Yu and Zhou, Ziqi and Wang, Xianlong and Zhang, Yanjun and Chen, Chao},  
booktitle={Proceedings of the Thirty-Third International Joint Conference on Artificial Intelligence (IJCAI'24)},
year={2024}
}

@inproceedings{xie2023badexpert,
  title={BaDExpert: Extracting Backdoor Functionality for Accurate Backdoor Input Detection},
  author={Xie, Tinghao and Qi, Xiangyu and He, Ping and Li, Yiming and Wang, Jiachen T and Mittal, Prateek},
  booktitle={Proceedings of the Twelfth International Conference on Learning Representations (ICLR'24)},
  year={2024}
}

@article{li2025robust,
  title={Robust Generative Adaptation Network for Open-Set Adversarial Defense},
  author={Li, Yanchun and Huang, Long and Tian, Shujuan and Liu, Haolin and Li, Zhetao},
  journal={IEEE Transactions on Information Forensics and Security},
  year={2025},
  publisher={IEEE}
}

@inproceedings{zhangdenial,   title={Denial-of-Service or Fine-Grained Control: Towards Flexible Model Poisoning Attacks on Federated Learning},   author={Zhang, Hangtao and Yao, Zeming and Zhang, Leo Yu and Hu, Shengshan and Chen, Chao and Liew, Alan and Li, Zhetao},   booktitle = {Proceedings of the 31st International Joint Conference on Artificial Intelligence, {IJCAI'23}},   year={2023}
}

@inproceedings{zhou2023advclip,
  title={Advclip: Downstream-agnostic adversarial examples in multimodal contrastive learning},
  author={Zhou, Ziqi and Hu, Shengshan and Li, Minghui and Zhang, Hangtao and Zhang, Yechao and Jin, Hai},
  booktitle={Proceedings of the 32nd ACM International Conference on Multimedia (MM'23)},
  pages={6311--6320},
  year={2023}
}

@article{zhang2024badrobot,
  title={BadRobot: Manipulating Embodied LLMs in the Physical World},
  author={Zhang, Hangtao and Zhu, Chenyu and Wang, Xianlong and Zhou, Ziqi and Yin, Changgan and Li, Minghui and Xue, Lulu and Wang, Yichen and Hu, Shengshan and Liu, Aishan and others},
  journal={arXiv preprint arXiv:2407.20242},
  year={2024}
}

@article{wang2024trojanrobot,
  title={TrojanRobot: Backdoor Attacks Against Robotic Manipulation in the Physical World},
  author={Wang, Xianlong and Pan, Hewen and Zhang, Hangtao and Li, Minghui and Hu, Shengshan and Zhou, Ziqi and Xue, Lulu and Guo, Peijin and Wang, Yichen and Wan, Wei and others},
  journal={arXiv preprint arXiv:2411.11683},
  year={2024}
}

@inproceedings{wangunlearnable,
  title={Unlearnable 3D Point Clouds: Class-wise Transformation Is All You Need},
  author={Wang, Xianlong and Li, Minghui and Liu, Wei and Zhang, Hangtao and Hu, Shengshan and Zhang, Yechao and Zhou, Ziqi and Jin, Hai},
  booktitle={The 38th Conference on Neural Information Processing Systems (NeurIPS'24)},
  year={2024}
}

@article{yao2024reverse,
  title={Reverse backdoor distillation: Towards online backdoor attack detection for deep neural network models},
  author={Yao, Zeming and Zhang, Hangtao and Guo, Yicheng and Tian, Xin and Peng, Wei and Zou, Yi and Zhang, Leo Yu and Chen, Chao},
  journal={IEEE Transactions on Dependable and Secure Computing},
  year={2024},
}

@article{song2024pb,
  title={PB-UAP: Hybrid Universal Adversarial Attack For Image Segmentation},
  author={Song, Yufei and Zhou, Ziqi and Li, Minghui and Wang, Xianlong and Zhang, Hangtao and Deng, Menghao and Wan, Wei and Hu, Shengshan and Zhang, Leo Yu},
  journal={arXiv preprint arXiv:2412.16651},
  year={2024}
}

@article{wang2024breaking,
  title={Breaking Barriers in Physical-World Adversarial Examples: Improving Robustness and Transferability via Robust Feature},
  author={Wang, Yichen and Chou, Yuxuan and Zhou, Ziqi and Zhang, Hangtao and Wan, Wei and Hu, Shengshan and Li, Minghui},
  journal={arXiv preprint arXiv:2412.16958},
  year={2024}
}
}


\end{document}